\documentclass[10pt,twocolumn,letterpaper]{article}

\usepackage{cvpr}              
\usepackage{times}  
\usepackage{graphicx} 

\usepackage{graphicx}
\usepackage{amsmath}
\usepackage{amssymb}
\usepackage{booktabs}
\usepackage{amsfonts}
\usepackage{booktabs}
\usepackage{cite}
\usepackage{algorithmic}
\usepackage{textcomp}
\usepackage{xcolor}
\usepackage{dsfont}
\usepackage{multirow}
\usepackage{newfloat}
\usepackage{listings}
\usepackage{diagbox}
\usepackage[accsupp]{axessibility}

\usepackage[pagebackref,breaklinks,colorlinks]{hyperref}

\usepackage[capitalize]{cleveref}
\crefname{section}{Sec.}{Secs.}
\Crefname{section}{Section}{Sections}
\Crefname{table}{Table}{Tables}
\crefname{table}{Tab.}{Tabs.}

\def\cvprPaperID{4068} 
\def\confName{CVPR}
\def\confYear{2023}

\begin{document}

\title{DPF: Learning \underline{D}ense \underline{P}rediction \underline{F}ields with Weak Supervision}

\author{Xiaoxue Chen$^{1}$, Yuhang Zheng$^2$,  Yupeng Zheng$^3$\\Qiang Zhou$^1$, Hao Zhao$^{1}$, Guyue Zhou$^{1}$, Ya-Qin Zhang$^1$ \\
	$^1$AIR, Tsinghua University $^2$BUAA $^3$CASIA\\
	{\tt\small \{chenxiaoxue, zhaohao\}@air.tsinghua.edu.cn},
 	{\tt\small zyh\_021@buaa.edu.cn}
  }
 
\maketitle

\begin{abstract}
Nowadays, many visual scene understanding problems are addressed by dense prediction networks. But pixel-wise dense annotations are very expensive (e.g., for scene parsing) or impossible (e.g., for intrinsic image decomposition), motivating us to leverage cheap point-level weak supervision. However, existing pointly-supervised methods still use the same architecture designed for full supervision. In stark contrast to them, we propose a new paradigm that makes predictions for \textbf{point coordinate queries}, as inspired by the recent success of implicit representations, like distance or radiance fields. As such, the method is named as dense prediction fields (DPFs). DPFs generate expressive intermediate features for continuous sub-pixel locations, thus allowing outputs of an arbitrary resolution. DPFs are naturally compatible with point-level supervision. We showcase the effectiveness of DPFs using two substantially different tasks: high-level semantic parsing and low-level intrinsic image decomposition. In these two cases, supervision comes in the form of single-point semantic category and two-point relative reflectance, respectively. As benchmarked by three large-scale public datasets PASCALContext, ADE20K and IIW, DPFs set new state-of-the-art performance on all of them with significant margins. Code can be accessed at \url{https://github.com/cxx226/DPF}.
\end{abstract}

\section{Introduction}
\label{sec:intro}

The field of visual scene understanding aims to recover various scene properties from input images, e.g., semantic labels \cite{gupta2015indoor}, depth values \cite{saxena2008make3d}\cite{zheng2023steps}, edge existence \cite{arbelaez2010contour} or action affordance \cite{chen2022cerberus}. Successful and comprehensive scene understanding is the cornerstone of various emerging artificial intelligence applications, like autonomous driving, intelligent robots or smart manufacturing. Albeit difficult, this field has seen great progress thanks to end-to-end dense prediction networks like DPT \cite{ranftl2021vision} and large-scale densely-labelled datasets like ADE20K \cite{zhou2019semantic}. If we can densely label every property that we care about, totally solving the visual scene understanding problem seems a matter of time.

\begin{figure}[t]
  \centering
  \includegraphics[width=1\linewidth]{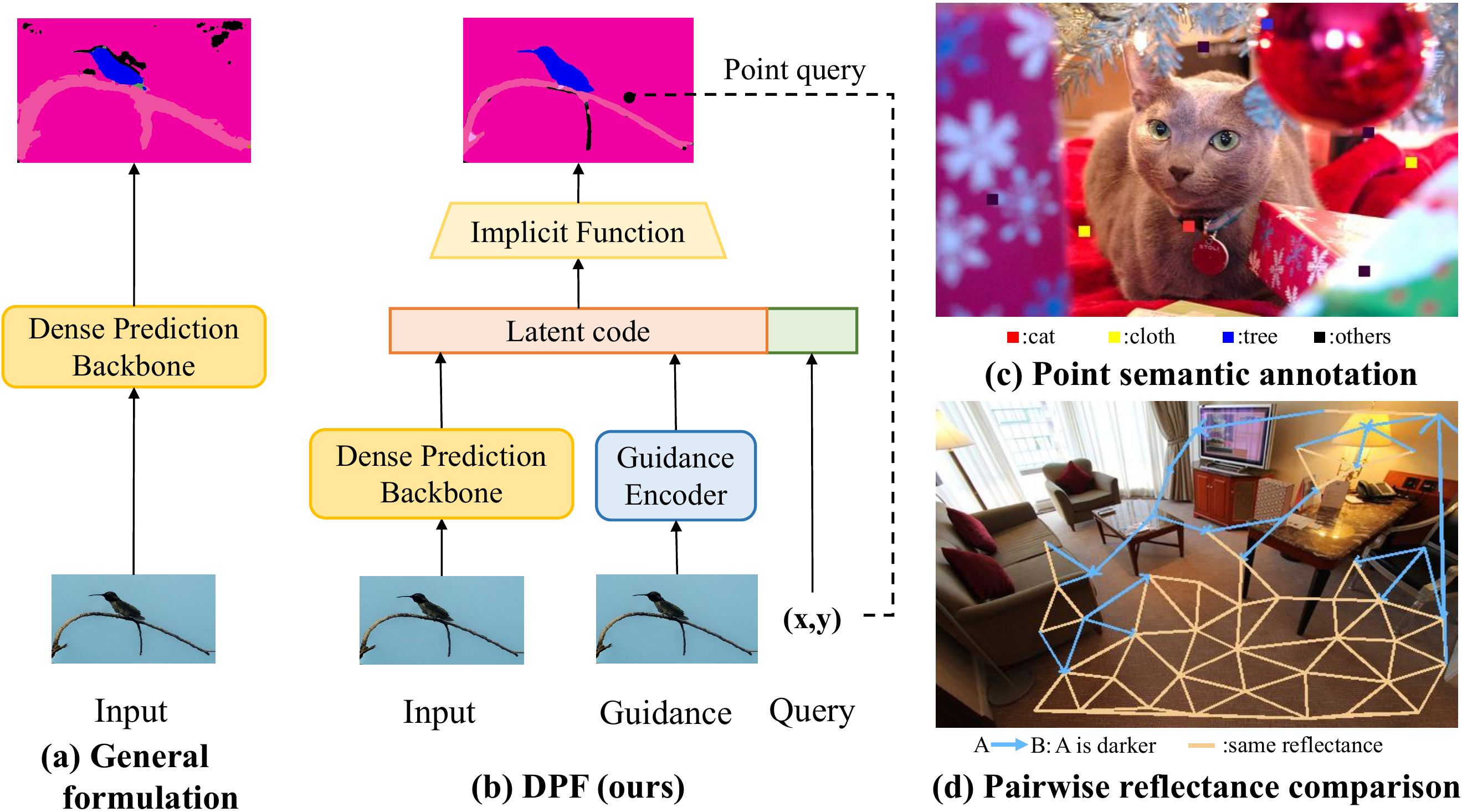}
  \caption{(a) Existing dense prediction formulation. (b) Our DPF formulation. (c) Semantic annotation for single points. (d) Pairwise reflectance annotation between two points.}
  \label{fig:teaser} 
\end{figure}

However, dense annotations are usually too expensive or impossible to obtain. According to the Cityscapes paper \cite{cordts2016cityscapes}, it takes 1.5 hours to generate a high-quality semantic annotation map for a single image. What's worse, for the problem of decomposing an image into reflectance and shading \footnote{Intrinsic image decomposition.}, it's impossible for humans to provide pixel-wise ground truth values. As such, the largest intrinsic image decomposition dataset IIW \cite{bell2014intrinsic} is annotated in the form of pair-wise reflectance comparison between two points. Annotators are guided to judge whether the reflectance of one point is darker than that of another point or not.

Given the importance of dense prediction and the difficulty of obtaining dense annotations, we focus on learning with point-level weak supervision. Fig.~\ref{fig:teaser}-c shows an example of point-level semantic scene parsing annotation. The sole red point on the cat is annotated as \emph{cat}, which is much more cheaper than delineating the cat's contours. Fig.~\ref{fig:teaser}-d shows human judgement of relative reflectance annotation between every pair of two points. Since the floor has a constant reflectance, point pairs on the floor are annotated with the \emph{equal} label. Since the table has a darker reflectance than the floor, pairs between the table point and the floor point are annotated with the \emph{darker} label. 

How could we effectively learn dense prediction models from these kinds of point-level weak supervision? To this end, existing pointly-supervised methods leverage unlabelled points using various techniques like online expansion \cite{qian2019weakly}, uncertainty mixture \cite{zhao2020pointly}\cite{tian2022vibus} or edge guidance \cite{fan2018revisiting}. But they all exploit conventional formulations shown in Fig.~\ref{fig:teaser}-a, by converting point-level supervision into dense ground truth maps with \emph{padded ignore values}. By contrast, we seek alternative network architectures that are naturally compatible with point-level supervision. Specifically, we take inspiration from the success of neural implicit representations. DeepSDF \cite{park2019deepsdf} takes 3D coordinates as input and predicts signed distance values. NeRF \cite{mildenhall2020nerf} takes 5D coordinates as input and predicts radiance/transparency values. Similarly, our method takes 2D coordinates as input and predicts semantic label or reflectance values, as shown in Fig.~\ref{fig:teaser}-b. An intriguing feature of this new scheme is that high-resolution images can be encoded as guidance in a natural way, because this new continuous formulation allows outputs of arbitrarily large or small resolution. Borrowing names from the research community of distance or radiance fields, our method is called \textbf{dense prediction fields (DPFs)}.

In order to show that DPF is a strong and generic method, we use two pointly-supervised tasks: semantic scene parsing and intrinsic image decomposition. These two tasks differ in many aspects: (1) Scene parsing is a high-level cognitive understanding task while intrinsic decomposition is a low-level physical understanding task; (2) Scene parsing outputs discrete probability vectors while intrinsic decomposition outputs continuous reflectance/shading values; (3) Scene parsing is annotated with single points while intrinsic decomposition is annotated with two-point pairs. Interestingly and surprisingly, our method achieves new state-of-the-art results on both of them, as benchmarked by three widely used datasets PASCALContext, ADE20K and IIW.

To summarize, the contributions of our work include:
\begin{itemize}
\item We propose a novel methodology for learning dense prediction models from point-level weak supervision, named DPF. DPF takes 2D coordinates as inputs and allows outputs of an arbitrary resolution.
\item We set new state-of-the-art performance on PASCALContext and ADE20K datasets for scene parsing and IIW dataset for intrinsic decomposition with point-level weak supervision. Codes are publicly available.
\item With systematic ablations, visualization and analysis, we delve into the mechanism of DPF and reveal that its superior performance is credited to locally smooth embeddings and high-resolution guidance.
\end{itemize}

\section{Related Work}
\label{sec:formatting}
\subsection{Intrinsic image decomposition}

Complete scene de-rendering \cite{karsch2011rendering,li2020inverse} is a long-term goal in visual intelligence, requiring many properties to be understood, like geometry\cite{ji2017surfacenet,xu2019unstructuredfusion}, room layout \cite{zhao2017physics,huang2018holistic,chen2022pq,gao2023semi}, lighting \cite{song2019neural,gardner2019deep}, and material \cite{meka2018lime,zheng2016deep,degol2016geometry}. \textbf{Intrinsic decomposition} is the minimal formulation that decomposes a natural image into reflectance and shading. Since the problem is severely ill-posed, conventional methods \cite{shen2008intrinsic,shen2011intrinsic,chen2013simple,grosse2009ground,shen2013intrinsic} resort to optimization algorithms with hand-crafted priors. Recently, many deep learning methods\cite{zoran2015learning,narihira2015learning,janner2017self,baslamisli2018joint,das2022intrinsic,das2022pie,baslamisli2018cnn,baslamisli2021shadingnet,forsyth2021intrinsic} have been proposed to solve it. \cite{ma2018single,liu2020unsupervised,zhang2021unsupervised} explore to address this problem in unsupervised manners. \cite{narihira2015direct,zhou2015learning} apply a CNN network to directly predict reflectance or shading. \cite{nestmeyer2017reflectance} develops a joint bilateral filtering method to leverage strong prior knowledge about reflectance constancy.  \cite{fan2018revisiting} adopts a guided, edge-preserving domain filter to generate realistic reflectance. \cite{li2018cgintrinsics} proposes a new end-to-end training pipeline that learns better decomposition by leveraging a large-scale synthetic dataset CGIntrinsics. \cite{li2020inverse,zhou2019glosh} introduce novel lighting representations to obtain a complete scene reconstruction including reflectance, shape, and lighting. IRISFormer \cite{zhu2022irisformer} adopts a transformer architecture to simultaneously estimate depths, normals, spatially-varying albedo, roughness and lighting from a single image. In this work, we focus on pointly-supervised intrinsic decomposition. Specifically, we benchmark on the IIW dataset \cite{bell2014intrinsic}, which is annotated with sparse, pairwise comparison labels. Although many of the above works are also evaluated on IIW, none of them are specifically designed for point supervision. Instead, our DPF method is \textbf{naturally compatible} with point supervision and achieves superior performance compared with all prior works.

\subsection{Scene parsing and weak supervision}

The goal of scene parsing is to classify all pixels in the image into corresponding categories. However, dense annotation for images, which costs a lot, is still critical to the success of scene parsing. This fact gives rise to the research of dense prediction with weak supervision. One line of works focuses on the usage of pseudo labels. Although prior methods of harvesting pseudo labels are designed in various manners, they rely on proper thresholds \cite{wei2016stc,wei2017object,bearman2016s,qian2019weakly,zhao2020pointly}. Among all, uncertainty mixture \cite{zhao2020pointly} that has the capacity of choosing the threshold automatically achieves strong results on the PASCALContext and ADE20K dataset. Recently, transformer based models have made great progress in scene parsing. The vision transformer (ViT) backbone \cite{dosovitskiy2020image} significantly benefits dense prediction due to its characteristics of maintaining a representation with constant spatial resolution throughout all processing stages and having a global receptive ﬁeld at every stage. Our method is based on the ViT backbone, leveraging the self-attention mechanism to better propagate supervision signals from sparse points to \textbf{all patch tokens}.

\subsection{Implicit neural representation}

Implicit neural representation is a paradigm that maps coordinates to signals in a specific domain with neural networks. On account of the continuous and differentiable deep implicit function, it can capture intricate details, bringing conspicuous performance in 3D reconstruction \cite{genova2019learning,michalkiewicz2019implicit,park2019deepsdf}. Recent researches also show the effectiveness of the implicit representation on 2D tasks. The Local Implicit Image Function \cite{chen2021learning} learns a continuous image representation that can be queried in arbitrary resolution. The Joint Implicit Image Function (JIIF) \cite{tang2021joint} formulates guided depth super-resolution as a neural implicit image interpolation problem. SIREN \cite{sitzmann2020implicit} leverages periodic activation functions for implicit neural representations, which are ideally suited for representing complex natural signals and their derivatives. The Implicit Feature Aliment Function \cite{hu2022learning} implicitly aligns the feature maps at different levels and is capable of producing segmentation maps in arbitrary resolutions. Based on the fact that point queries and point supervision are inherently compatible, we explore the employment of neural implicit image interpolation with point queries under weak supervision. 

\section{Method}

\begin{figure*}[t]
  \centering
  \includegraphics[width=0.8\linewidth]{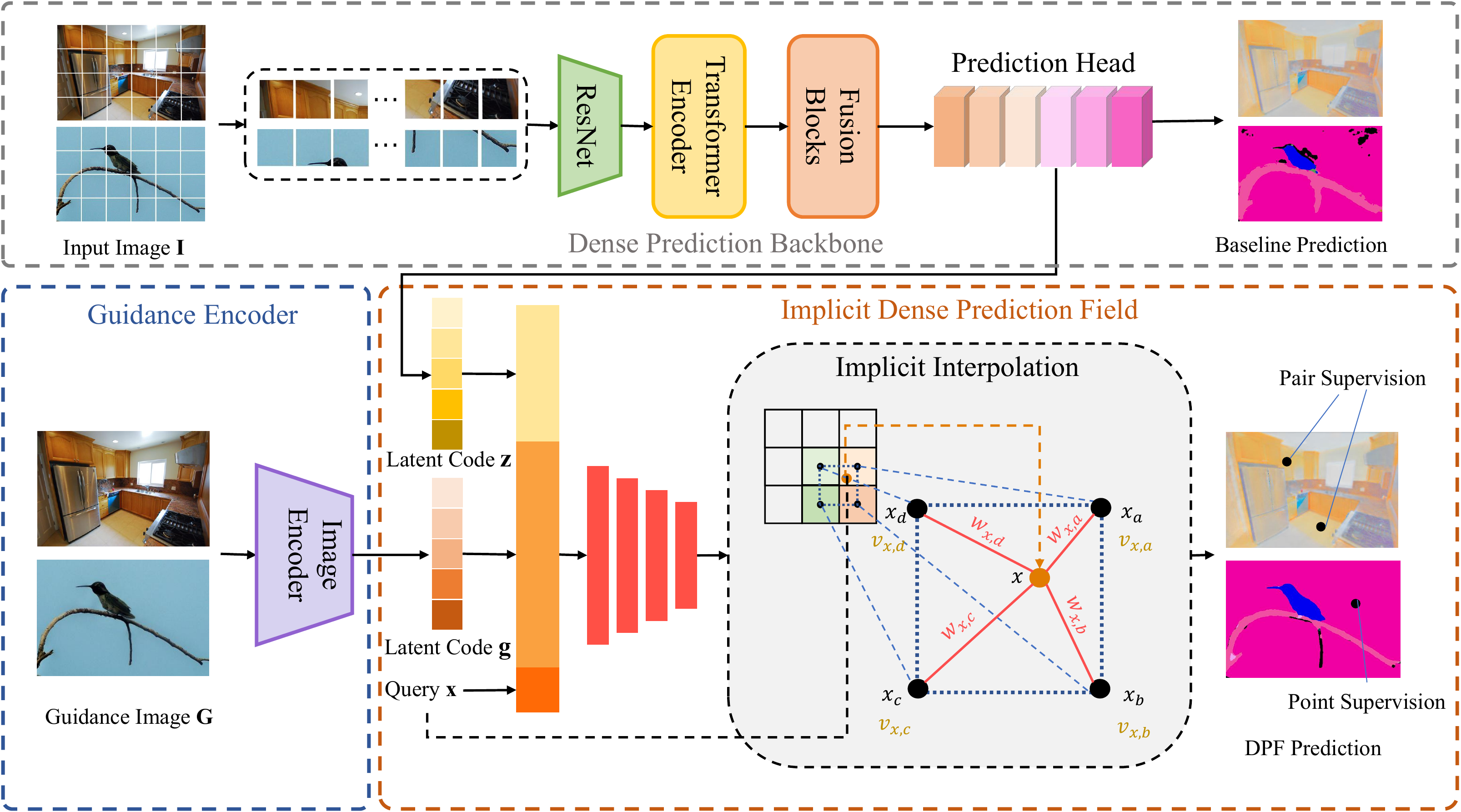}
  \caption{\textbf{Overall architecture}. Our model consists of three components: a dense prediction backbone to extract high-level features and make baseline predictions, a guidance encoder to encode guidance features, and an implicit dense prediction field to make predictions at point coordinate queries. The upper figures are for intrinsic decomposition while the lower figures are for scene parsing.}
  \label{fig:main} 
\end{figure*}

\subsection{Dense Prediction with Point Supervision}

Given an input image, dense prediction is the task of predicting an entity of interest (a label or a real number) for each pixel in the image. Previous works \cite{ranftl2021vision}\cite{chen2022cerberus} usually use pixel-wise annotations as the supervision to train dense prediction models. However, it is time-consuming to annotate in a pixel-wise manner. Sometimes it's \textbf{even impossible} to annotate the pixel with a certain value, for example, annotating an in-the-wild image with specific reflectance. Therefore, in this work, we focus on dense prediction with point supervision and propose a novel neural network to resolve it. Specifically, we introduce a dense prediction field (DPF) that predicts a corresponding value for each continuous 2D point on the imaging plane. Moreover, inspired by the recent success of implicit representations \cite{park2019deepsdf}\cite{li2023lode}, we use an implicit neural function to implement the DPFs. Mathematically, given a coordinate query $\rm x$ in the image,
\begin{align}
  \rm DPF(x) = v_x, \quad v_x \in \mathbb{R}^c, 
\end{align}
where $\rm c$ is the dimension of predicted entity. Due to its continuous nature, DPF is spatially consistent thus can achieve superior performance under sparse point supervision. 

To verify the effectiveness of the proposed dense prediction field, we benchmark it on two different types of pointly-supervised datasets: (1) datasets with sparsely labeled semantic category information like PASCALContext and ADE20K, and (2) datasets labeled with sparse pairwise comparisons like IIW. Both PASCALContext and ADE20K are designed for semantic parsing. While IIW is aimed at decomposing natural images into intrinsic reflectance and shading. These two types of datasets involve different prediction targets and different losses. Experiments show that we achieve SOTA performance on all three datasets which demonstrates the \textbf{generalizability} of our method.

\subsection{Network Architecture}

As depicted in Fig. \ref{fig:main}, our network is composed of three components: a dense prediction backbone $\rm h_\lambda$, a guidance encoder $\rm g_\eta$, and an implicit dense prediction field $\rm f_\theta$. The overall formulation of DPF is:
\begin{align}
\rm v_x = f_\theta(z,g,x), \label{eq:formutation}
\end{align}
where z and g are the latent codes extracted from $\rm h_\lambda$ and $\rm g_\eta$ respectively and $\rm x$ is the point query coordinate.

\textbf{Dense prediction backbone.} Previous works\cite{tang2021joint} typically formulate an image-based implicit field into:  
\begin{align}
\rm v_x = f_\theta(E(I),x)
\end{align}
where $\rm E$ is an encoder network to extract low-level visual features as  latent code. However, considering the importance of high-level semantic information extracted by specially designed dense prediction networks, we propose a novel paradigm that combines a dense prediction backbone and an implicit field. Specifically, given an input image I, we first feed the image into the dense prediction backbone:
\begin{align}
  \rm V, z = h_\lambda\left( I \right),\quad V = h_{\lambda1}(z) \label{eq:backbone}
\end{align}
$\rm h_{\lambda1}$ is the prediction head of $\rm h_{\lambda}$, $\rm V$ is the baseline dense prediction value and z is the high-level features extracted from the output of the intermediate layer of $\rm h_\lambda$, specifically z is the output of the penultimate layer (before mapping the number of feature channels to the dimensionality of prediction targets).  We impose a loss on V as \textbf{auxiliary supervision}, which provides constraints on the predicted value ($v$ in the later Eq. \ref{interpolation}) of the implicit field while facilitating the latent code z to acquire corresponding high-level information. This design is ablated in Tab. \ref{tab:guide}.


The paradigm in Eq. \ref{eq:formutation} can be applied on top of any plug-and-play dense prediction models. To verify this, we choose a CNN-based network FastFCN\cite{wu2019fastfcn} and a ViT-based network DPT\cite{ranftl2021vision} as the backbones.

\textbf{Guidance encoder.} Guided image filtering \cite{he2012guided} is an effective edge-preserving smoothing operator based on a guidance image. Following previous works\cite{fan2018revisiting}, we also introduce an extra guidance image $G$. We believe the content of the guidance image can benefit the learning of interpolation parameters (described in Sec 3.3) and make the DPF outputs better aligned with the high-resolution guidance image. We directly use the input image of different resolutions as the guidance image instead of introducing a task-specific guidance map (e.g., the edge guidance in \cite{fan2018revisiting,das2022pie}) that requires domain-specific pre-processing. 

We use an EDSR \cite{lim2017enhanced} network as the guidance encoder, and extract features from the guidance image:
\begin{align}
 \rm g = g_\eta\left( G \right)
\end{align}

$\rm g$ also serves as a latent code, but it contains low-level local features which are complementary to z. The EDSR model consists of 16 residual blocks without upsampling layers, and we use the output of the last block as g. Both the two latent codes provide important information to support the learning of DPFs. Their effects and differences will be shown in Fig. \ref{fig:tsne}. In the following section, we will describe our implicit dense prediction field in detail. 

\subsection{Implicit Dense Prediction Field}

Given the coordinate x of a point on the image plane, we are aiming to query its value $v_x$ in the dense prediction field. Notably, x can be a random coordinate value sampled from a continuous space, so we can't directly extract the corresponding value from a discrete dense prediction map. A straightforward way to get  $v_x$  is to interpolate the dense prediction values of neighbor pixels, as illustrated in Fig. \ref{fig:main} (Implicit Interpolation). Specifically, the corresponding dense prediction value $v_x$ is defined as:

\begin{align}
 v_x = \sum_{i \in N_x} w_{x,i} v_{x,i}, \quad \sum_{i \in N_x} w_{x,i} = 1 \label{interpolation}
\end{align}

where $N_x$ is the set of neighbor pixels of x, $v_{x,i}$ is the dense prediction value of pixel $\rm i$, $w_{x,i}$ is the interpolation weight between $\rm x$ and $\rm i$. For the scene parsing tasks with multiple semantic categories, the values are vectors of length $\rm c$, where $\rm c$ is the number of categories. For the reflectance prediction, the values are scalars. In practice, all the coordinates are normalized into $(-1,1)$ with the image center as the origin. This normalization step allows us to conveniently combine latent codes of different resolutions ($\rm g$ and $\rm z$ specifically).

Inspired by deep implicit function methods \cite{park2019deepsdf,tang2021joint,chen2021learning}, we use a deep neural network to get the interpolation weights and dense prediction values. Given the input image feature z and the guidance feature g, we leverage an MLP to learn the interpolation weights and values between coordinate $\rm x$ and its neighbor pixel $\rm i$:
\begin{align}
     \hat{w}_{x,i}, v_{x,i} = {\rm MLP (z_i,g_i, \gamma (\Delta x ) ) } \label{mlp}
\end{align}
where $\rm z_i, g_i$ is the corresponding latent code of pixel $\rm i$ that is extracted from z and g. $\rm \Delta x = x_i -x$ is a relative coordinate, and $\rm x_i$ is the coordinate of $\rm i$. This relative coordinate indicates the spatial affinity between query point $\rm x$ and its neighbor pixel $\rm i$. Furthermore, we also apply a positional encoding $\gamma(\cdot)$ following \cite{mildenhall2020nerf} to leverage higher frequency spatial signals:
\begin{align}
 \gamma(x) = (sin(2^0\pi x), cos(2^0\pi x),...,sin(2^{l}\pi x), cos(2^{l}\pi x))
\end{align}
 
In practice, we set $l= 9$. After Eq. \ref{mlp}, the interpolation weights are normalized through a softmax layer:
\begin{align}
 w_{x,i} = \frac{ {\rm exp}(\hat{w}_{x,i})}{\sum_{j \in N_x} {\rm exp}(\hat{w}_{x,j})} \label{normalize}
\end{align}
By integrating the interpolation (Eq. \ref{interpolation}) and the calculation of weights and values (Eq. \ref{mlp}, \ref{normalize}), the formulation of our implicit dense prediction field  can be represented as: 

\begin{align}
 v_x = {\rm f_\theta(z,g,x)}
\end{align}

where $\theta$ is the network parameters.


\subsection{Training Loss}

To get the prediction of DPFs, we use the coordinate of every pixel in the guide image as queries, and generate a prediction map of the same resolution of the guide image. For the scene parsing task, the number of channels of the prediction map is $\rm c$, where $\rm c$ is the number of semantic categories. As for intrinsic decomposition, the number of channels is 1. We supervise both the predictions of the dense prediction backbone and DPF using the same kind of loss functions. Specifically, we use a $\rm c$-way cross-entropy loss for scene parsing datasets. 

For the pairwise comparison dataset IIW, there are no absolute ground truth labels available. Instead, given the $\rm k$-th pair of comparison points $	\left\{k_1,k_2 	\right\}$, the relative reflectance annotation $J_k$ is classified into three labels: 
\begin{align}
J_k = 
\begin{cases}
1& \rm if\ k_1\ is\ \emph{darker}\ than\ k_2, \\
2&  \rm if\ k_1\ is\ \emph{lighter}\ than\ k_2, \\
E& \rm if\ reflectance\ of\ k_1\ and\ k_2\ are\ \emph{equal}.
\end{cases}
\end{align}
We denote the predicted reflectance of point $k_1$ and $k_2$ as $R_{k_1}$ and $R_{k_2}$, respectively. We use a standard SVM hinge loss to supervise the pairwise comparison data:
\begin{align}
 \mathcal{L}_{k} = \begin{cases}
max(0,\frac{R_{k_1}}{R_{k_2}}-\frac{1}{1+\delta+\epsilon}) & \rm if\ J_k = 1, \\
max(0,1+\delta+\epsilon-\frac{R_{k_1}}{R_{k_2}}) & \rm if\ J_k = 2,\\
max(0,
\begin{cases}
\frac{1}{1+\delta-\epsilon} - \frac{R_{k_1}}{R_{k_2}},\\
\frac{R_{k_1}}{R_{k_2}}-(1+\delta-\epsilon) 
\end{cases}) & \rm if\ J_k = E.
\end{cases}
\end{align}
$\epsilon$ and $\delta$ are hyper-parameters, and we set $\epsilon=0.08$ and $\delta = 0.12$ during training.

The total loss for all comparison pairs is defined as:
\begin{align}
\mathcal{L}_{\rm pairs} = \sum_{k \in{\rm P}} s_k \cdot \mathcal{L}_{k}
\end{align}
where P is the index set of all comparison pairs, and $s_k$ is the confidence score of each annotation provided by the dataset.
\section{Experiment}
\subsection{Datasets and Evaluation Protocols}

\textbf{Intrinsic decomposition.} We report results on the IIW dataset \cite{bell2014intrinsic}. The IIW dataset contains 5,230 indoor scene images, and 872,151 relative reflectance comparison pairs in total. Following the setting of \cite{fan2018revisiting}, we sort the IIW dataset by image ID, and put the first of every five images into the test set, and the rest into the training set. We employ weighted human disagreement rate (WHDR) as the evaluation metric. The classification of predicted reflectance comparison pairs can be calculated as:
\begin{align}
\hat{J}_k = 
\begin{cases}
1& \rm if\ \frac{R_{k2}}{R_{k1}} \textgreater 1 + \delta, \\
2&  \rm if\ \frac{R_{k1}}{R_{k2}} \textgreater 1 + \delta, \\
E& otherwise.
\end{cases}
\end{align}
where $\delta$ is the threshold to filter out negligible relative difference, which we set as 0.1 in the evaluation. The WHDR is the error rate of $\hat{J}_k$ when compared with ${J}_k$.

\textbf{Scene parsing.} We benchmark DPFs on two scene parsing datasets: PASCALContext \cite{mottaghi2014role} and ADE20K \cite{zhou2019semantic}. For PASCALContext, 4998 samples are used for training and 5105 samples are used for testing. For ADE20K, 20210 images are used for training and 2000 images are used for testing. PASCALContext has 60 different semantic labels, and ADE20K has 150 different semantic labels. For fair comparison, we use the same point annotations as \cite{zhao2020pointly} uses. We choose the mean intersection over union (mIoU) score as the evaluation metric for both datasets.


\subsection{Comparisons with SOTA methods}

\begin{figure}[h]
  \centering
  \includegraphics[width=0.8\linewidth]{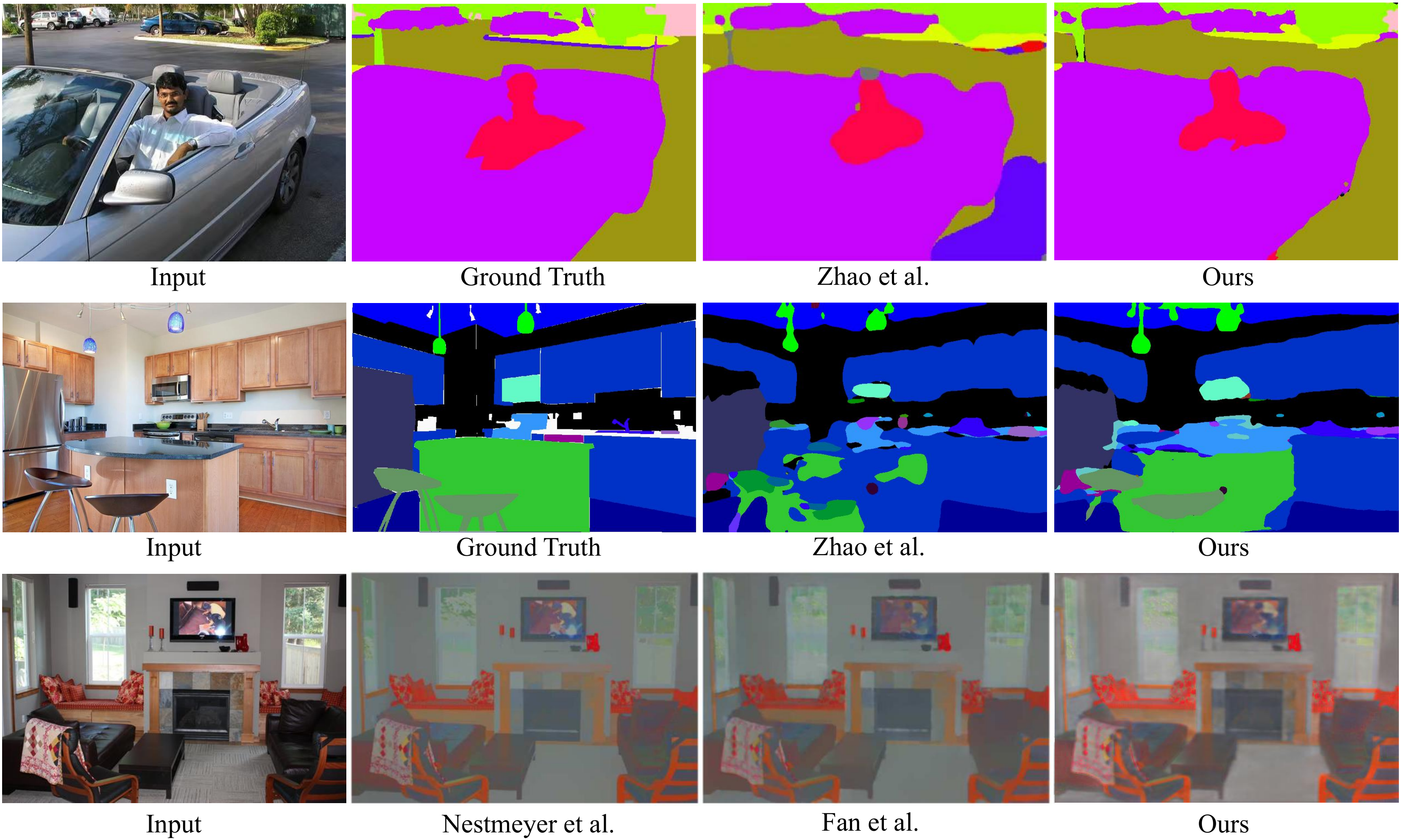}
  \caption{Qualitative comparisons on the PASCALContext (first row), ADE20K (second row) and IIW (last row), respectively.}
  \label{fig:sota} 
\end{figure}

\textbf{Intrinsic decomposition}. We provide the quantitative results of DPF (ViT based) on IIW in Tab. \ref{tab:iiw}. Our model outperforms the previous state-of-the-arts. Specifically, we achieve a 0.1\% boost over IRISformer \cite{zhu2022irisformer}, which introduces the OpenRooms (OR) dataset \cite{li2020openrooms} during training. Notably, many methods in Tab.\ref{tab:iiw} introduce synthetic datasets with full intrinsics ground truth, while our method is only trained on the pointly-annotated IIW dataset. Compared with the previous SOTA \cite{fan2018revisiting} trained only on IIW, our method promotes WHDR by 2.6\%, suggesting the effectiveness of our formulation using pairwise point comparison data. Fig. \ref{fig:sota} presents qualitative comparisons on three datasets, demonstrating the superior performance of DPFs compared with prior works. 

\begin{table}
	\centering
	\begin{tabular}{ccc}
		\toprule
		\textbf{Method} & \textbf{Training set} & \textbf{WHDR (\%) $\downarrow$} \\
		\midrule
        Sengupta et al. \cite{sengupta2019neural} &  CGP+IIW &  16.8 \\
		Li and Snavely\cite{li2018cgintrinsics} & CGI+IIW* & 16.2 \\
        Li et al.\cite{li2020inverse} & CGM+IIW & 15.9 \\
        Zhu et al.\cite{zhu2022irisformer} & OR+IIW & 12.0 \\
        \midrule
        Bell et al.\cite{bell2014intrinsic} & IIW & 20.6\\
         Nestmeyer et al.\cite{nestmeyer2017reflectance} & IIW & 17.7\\
        Bi et al.\cite{bi20151} & IIW & 17.7 \\
        Fan et al.\cite{fan2018revisiting} & IIW & 14.5 \\
        \midrule
		Ours & IIW &  \textbf{11.9} (+2.6) \\
		\bottomrule
	\end{tabular}
	\caption{Quantitative results on IIW.  Lower WHDR is better. IIW* indicates augmented IIW comparisons. CGI \cite{li2018cgintrinsics}, CGM \cite{li2020inverse}, CGP \cite{sengupta2019neural}, OR \cite{li2020openrooms} are all intrinsic decomposition datasets with dense labels. }
	\label{tab:iiw}
\end{table}

\textbf{Scene parsing}. Tab. \ref{tab:pascal},\ref{tab:ADE20K} provide the performance of ViT-based DPFs on PASCALContext and ADE20K respectively. For PASCALContext, the mIoU is significantly promoted from 36.1\% to 45.3\%. For ADE20K, the mIoU utperforms the previous SOTA by 5.0\%. 
This shows that our model also performs well under the supervision of single sparse labels. On the one hand, this is credited to the attention-based backbone (DPT) we use, which has already shown strong performance in the field of dense prediction tasks due to its global receptive field; on the other hand, our proposed DPF refines the dense prediction results with an implicit neural representation, naturally enabling smoother results under point supervision.

\begin{table}
	\centering
	\begin{tabular}{@{}lc@{}}
		\toprule
		\textbf{Method} & \textbf{mIoU (\%) $\uparrow$} \\
		\midrule
		Qian et al.\cite{qian2019weakly} w/o Online Ext& \text{29.70}\\
		Qian et al.\cite{qian2019weakly} w/ Online Ext& \text{30.00}\\
        Zhao et al.\cite{zhao2020pointly} w/o rGMM & \text{33.54}\\
        Zhao et al.\cite{zhao2020pointly} w/ rGMM & \text{36.07}\\
		\midrule
		Ours & \textbf{45.31} (+9.2) \\
		\bottomrule
	\end{tabular}
	\caption{Quantitative results on PASCALContext.}
	\label{tab:pascal}
\end{table}

\begin{table}
	\centering
	\begin{tabular}{@{}lc@{}}
		\toprule
		\textbf{Method} & \textbf{mIoU (\%) $\uparrow$} \\
		\midrule
		Qian et al.\cite{qian2019weakly} w/o Online Ext& \text{19.00}\\
		Qian et al.\cite{qian2019weakly} w/ Online Ext& \text{19.60}\\
        Zhao et al.\cite{zhao2020pointly} w/o rGMM & \text{26.33}\\
        Zhao et al.\cite{zhao2020pointly} w/ rGMM & \text{28.79}\\ 
		\midrule
		Ours & \textbf{33.84} (+5.0) \\
		\bottomrule
	\end{tabular}
	\caption{Quantitative results on ADE20K.}
	\label{tab:ADE20K}
\end{table}


\begin{figure}[h]
  \centering
  \includegraphics[width=0.8\linewidth]{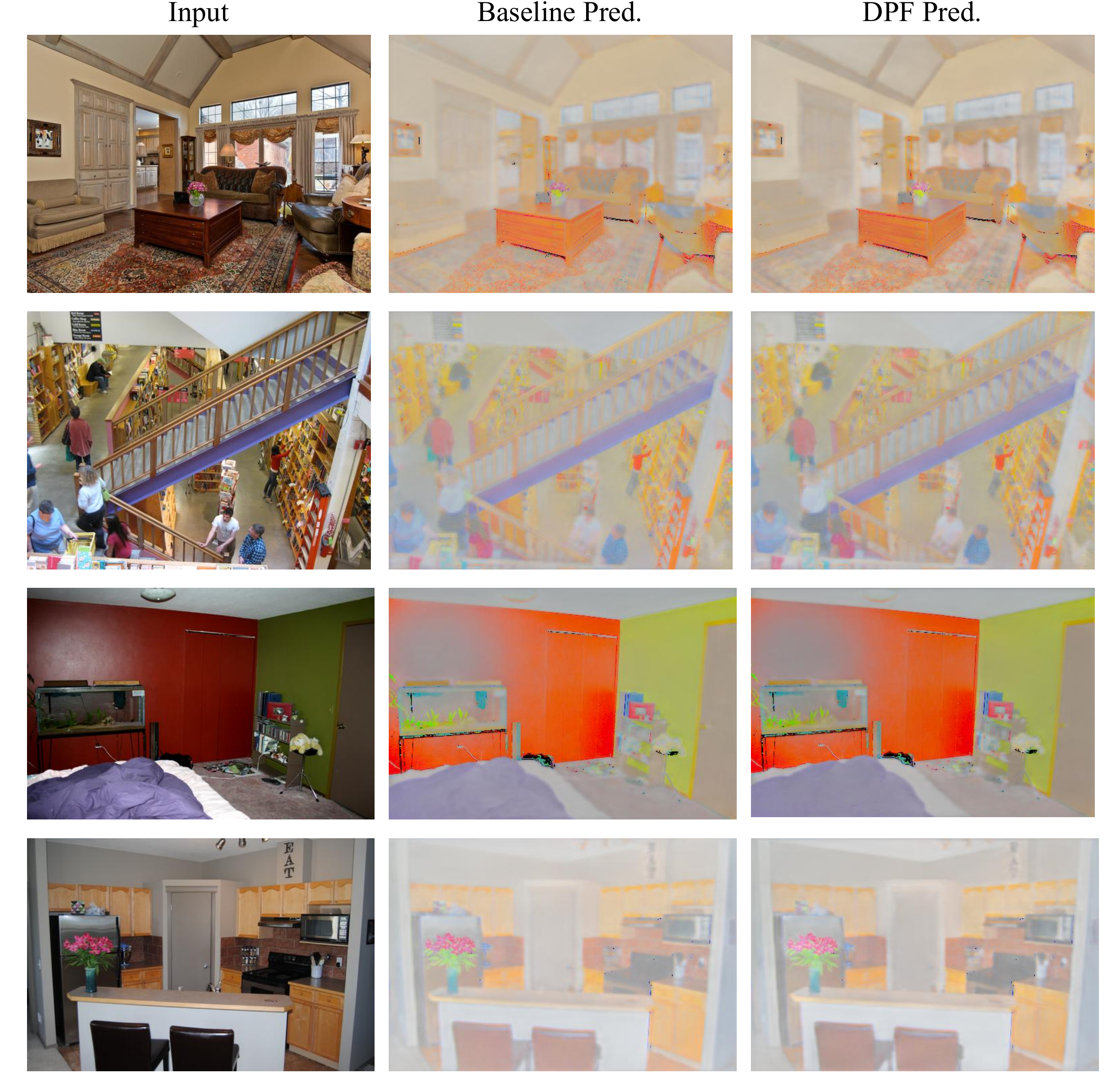}
  \caption{Qualitative prediction results on IIW.}
  \label{fig:iiw} 
\end{figure}

\begin{figure}[h]
  \centering
  \includegraphics[width=0.80\linewidth]{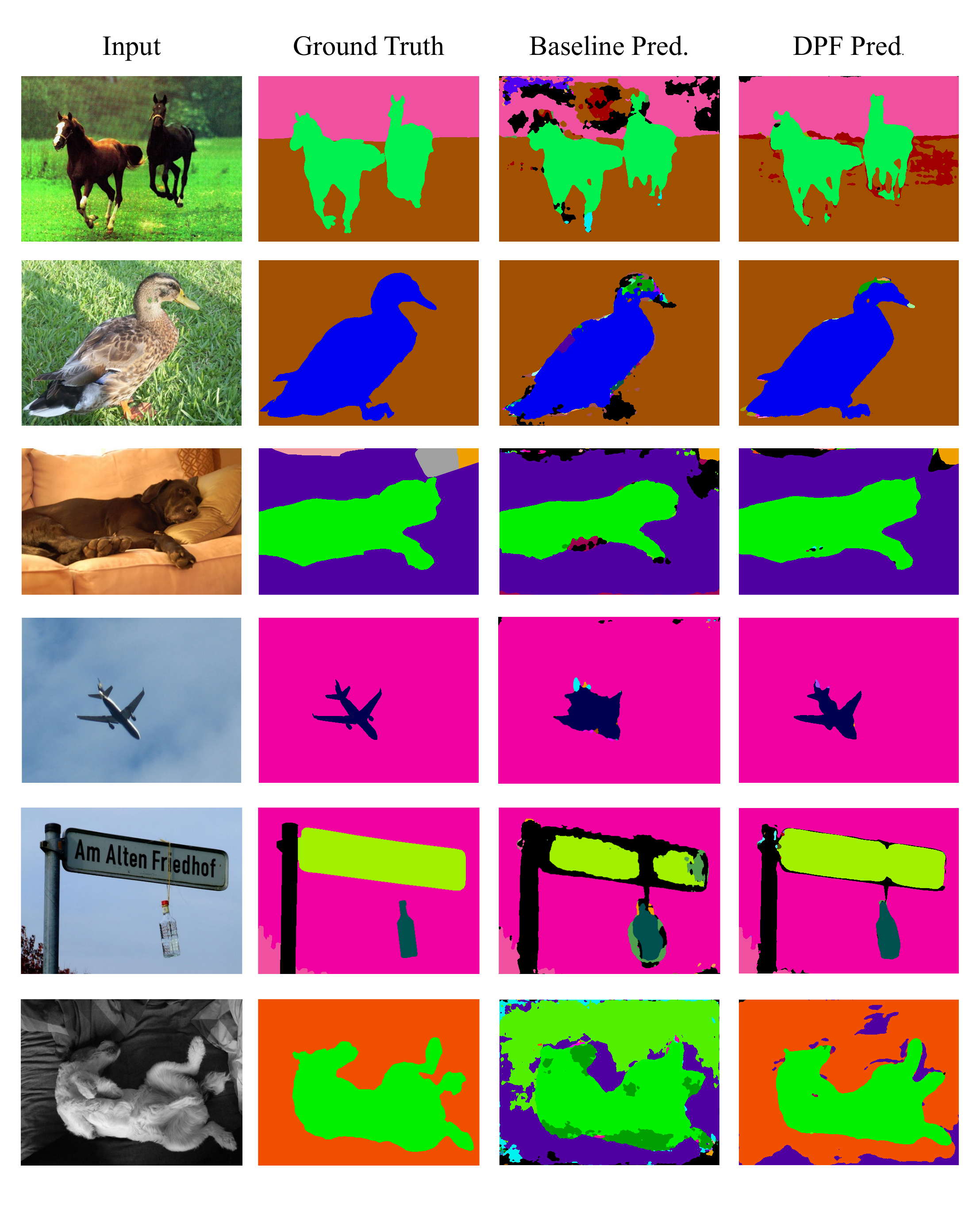}
  \caption{Qualitative prediction results on Pascal. Different colors represent different semantic categories.}
  \label{fig:Pascal} 
\end{figure}

\begin{figure}[h]
  \centering
  \includegraphics[width=0.80\linewidth]{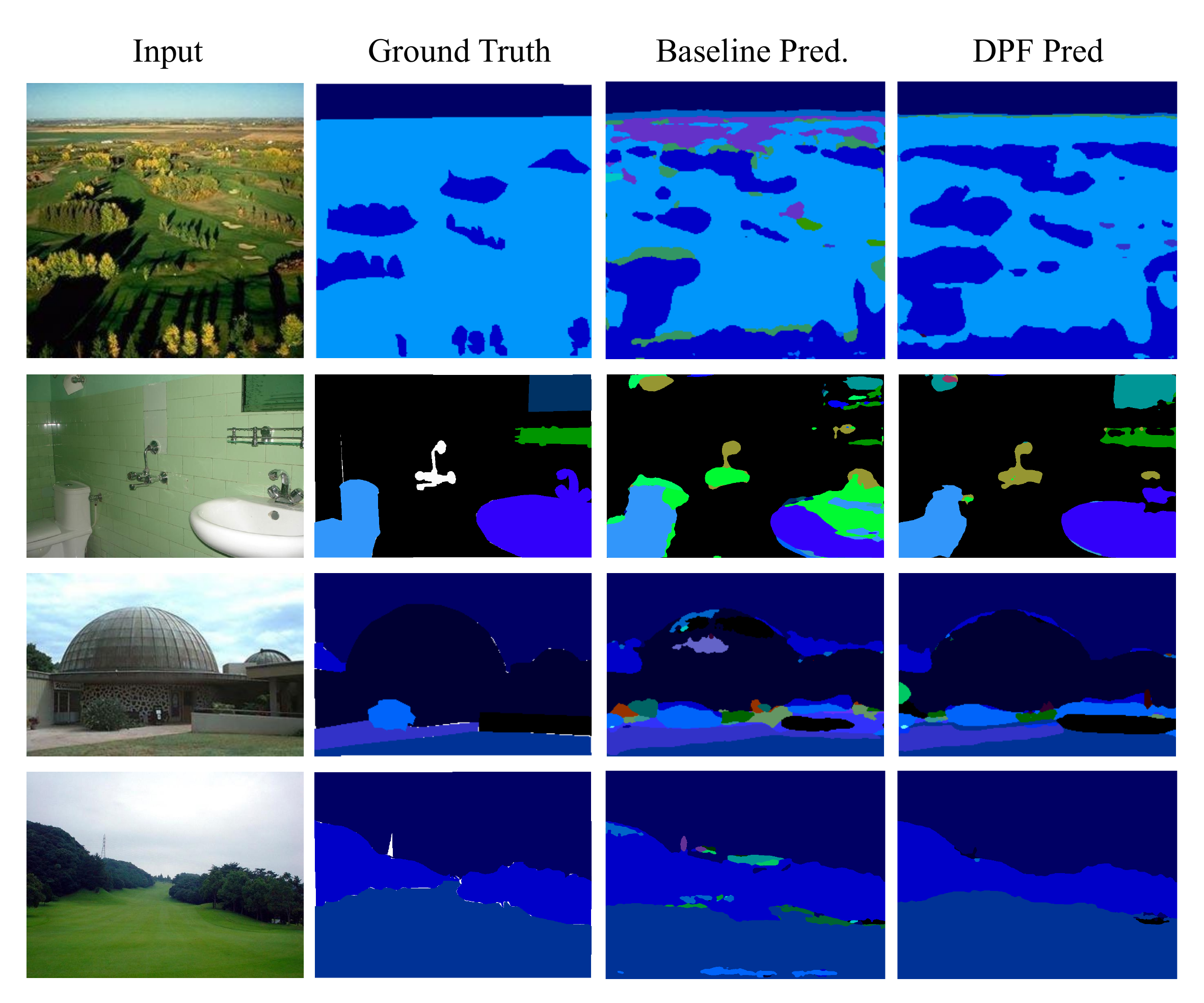}
  \caption{Qualitative prediction results on ADE20K.}
  \label{fig:ADE20K} 
\end{figure}


\subsection{Effectiveness of DPF on different backbones}
\begin{figure}[h]
  \centering
  \includegraphics[width=0.75\linewidth]{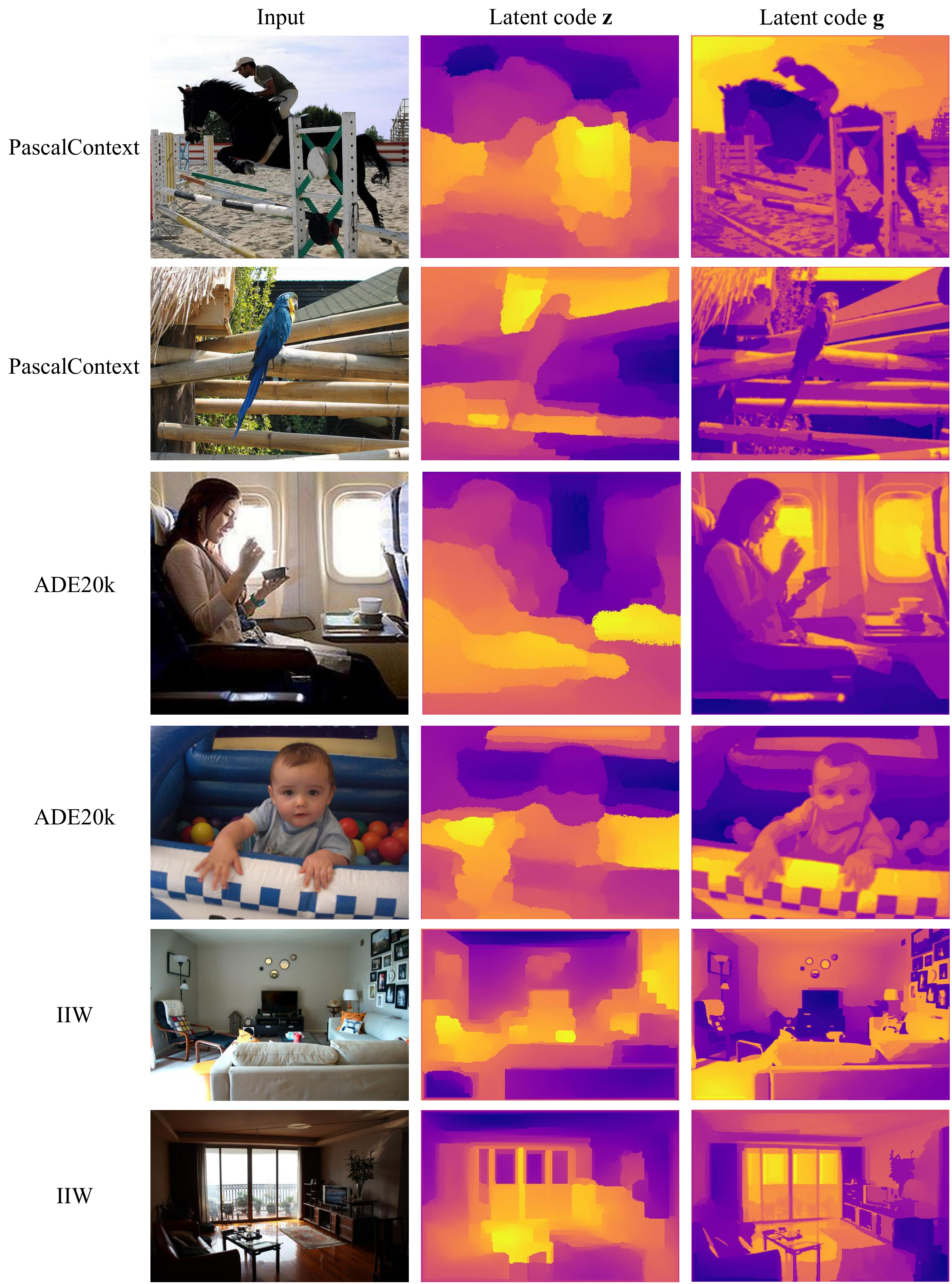}
  \caption{Visualization of t-SNE of latent codes.}
  \label{fig:tsne} 
\end{figure}

To further prove the effectivness of DPF, we train the CNN baseline (FastFCN), ViT baseline (DPT), and DPF with different backbones on all three datasets, and the results are shown in Tab. \ref{tab:dpf}. On all datasets, DPFs outperform the baselines significantly. Specifically, for ViT-based DPF, the mIoU is increased by 4.9\% on PASCALContext, performance on ADE20K increases by 3.4\%, and WHDR of IIW is decreased by 2.1\%, which indicate that DPF conclusively improves pointly-supervised dense prediction. This is credited to the representation of the latent codes, which combines high-level image features, low-level guidance features, and spatial information from relative query coordinates. Meanwhile, the implicit interpolation weights the values of neighbor pixels adaptively, making the dense prediction results more consistent. In addition,  using a transformer backbone leads to larger performance improvement than the CNN backbone due to the self-attention mechanism, as it can naturally help the propagation of sparse supervision with global patch interaction.

\begin{figure}[h]
  \centering
  \includegraphics[width=0.63\linewidth]{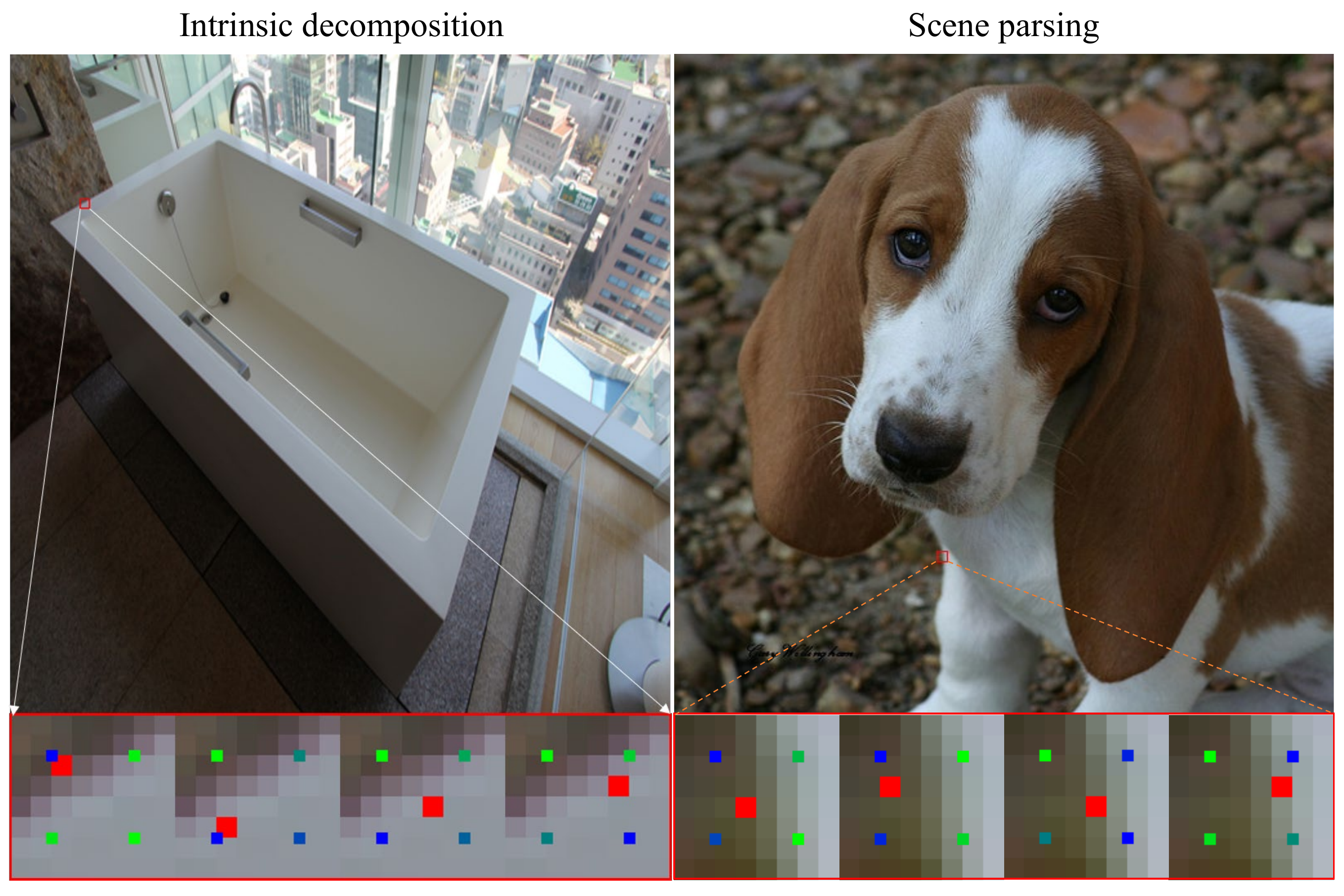}
  \caption{Visualization of the learned interpolation weights. }
  \label{fig:weights} 
\end{figure}

\textbf{Weight visualization}. Fig. \ref{fig:weights}  provides the visualization of the learned interpolation weights. The query pixel is in red, and the four corner pixels’ color indicates the learned interpolation weights. Higher weights are in bluer color, while lower weights are greener. It shows that DPF can successfully learn the interpolation weights depending on the location of the query point. When the query point shares the same reflectance or semantic label with its neighbor pixel, the weights will be higher. Conversely, the weights will be lower. We note that this kind of interpretable weight are learned through sparse annotation. This makes the DPF's prediction smoother and more accurate, while respecting the edges in input images.

\textbf{Qualitative results}. We provide visualization results on three datasets in Fig. \ref{fig:iiw}, \ref{fig:Pascal} and \ref{fig:ADE20K}, respectively. Fig. \ref{fig:Pascal}, \ref{fig:ADE20K} provides the results on scene parsing. Compared with the baseline prediction, DPF produces more accurate results. Specifically, as shown in the figures, there are a lot of noise patches in the baseline predictions, and these patches are misclassified, making the visualization results look very cluttered. Different from baseline, there are fewer misclassified patches in the DPF predictions. Besides, the predictions on the edge of objects are smoother for DPF, like the edge of the road sign on the 5th row in Fig. \ref{fig:Pascal}. Meanwhile, the segmentation of objects is also more precise. Take the airplane on the 4th row (Fig.\ref{fig:Pascal}) for an example, the shape of the airplane in DPF predictions is more reasonable, while the baseline result is relatively blurry. Fig. \ref{fig:iiw} presents qualitative results on IIW. As shown in the image, the prediction of DPF is smoother compared with the baseline prediction. For the purple quilt in the third row of Fig. \ref{fig:iiw}, DPF can distinguish reflections and wrinkles, and decompose the quilt into the same reflectance, while baseline prediction is not as flattened as ours. These results illustrate the capability of DPF on intrinsic decomposition.


\begin{table}
	\centering
	\begin{tabular}{c|ccc}
		\toprule
		\textbf{Datasets} & PASCAL & ADE20K & IIW \\
		\midrule
		CNN baseline & 37.3 & 26.0 & 17.9  \\
        DPF (CNN)  & \textbf{38.7} (+1.4) & \textbf{27.2} (+1.2) &  \textbf{17.2} (+0.7) \\
        \midrule
        ViT Baseline & 40.4 & 30.4 & 14.0  \\
        DPF (ViT) & \textbf{45.3} (+4.9)& \textbf{33.8} (+3.4) & \textbf{11.9} (+2.1) \\
		\bottomrule
	\end{tabular}
	\caption{DPF Quantitative results using different backbones.}
	\label{tab:dpf}
\end{table}


\begin{table}
	\centering
	\begin{tabular}{c|cc|c}
		\toprule
		\textbf{Datasets} & \textbf{w/o auxiliary} & \textbf{w/o guide} & \textbf{All} \\
		\midrule
 		PASCAL & 40.0 (-5.3) & 44.5 (-0.8)  &  \textbf{45.3}\\
        ADE20K & 28.3 (-5.5) & 32.9 (-0.9) & \textbf{33.8}  \\
        IIW & 25.2 (-13.3) & 12.5 (-0.6) &  \textbf{11.9} \\
		\bottomrule
	\end{tabular}
	\caption{Quantitative results on the effect of auxiliary supervision and guidance image.}
	\label{tab:guide}
\end{table}

\begin{table}[b]
    \small
	\centering
	\begin{tabular}{c|c|ccc}
		\toprule
		\textbf{Dataset} & \diagbox{\textbf{Guide}}{\textbf{Input}} & \textbf{128} & \textbf{256} & \textbf{512} \\
		\midrule
        \multirow{4}{*}{PASCAL}  
        & / & 13.7 & 27.2 & 40.4\\
        & 128 & 31.1 & - & -\\
		& 256  & 31.8 & 42.3 & -\\
        & 512  & \textbf{32.3} & \textbf{42.6} & \textbf{45.3}\\
        \midrule
        \multirow{4}{*}{ADE20K}  
        & / & 8.9 & 21.5 & 30.4\\
        & 128 & 15.9 & - & -\\
		& 256 & 16.4 & 29.0 & -\\
        & 512 & \textbf{17.1} & \textbf{29.2} & \textbf{33.8}\\
        \midrule
        \multirow{4}{*}{IIW}  
        & / & 22.2 & 17.9 & 14.0\\
        & 128 & 21.4 & - &  -\\
		& 256  & 21.0 &  16.5 & -\\
        & 512  & \textbf{20.6}  & \textbf{15.3}  & \textbf{11.9}\\
\bottomrule
	\end{tabular}
	\caption{Quantitative results with different input image resolutions and guidance resolutions.}
	\label{tab:resolution}
\end{table}

\subsection{Experiments on Network Architecture}

\textbf{Auxiliary supervision.} Tab. \ref{tab:guide} investigates the effects of auxiliary supervision on the dense prediction backbone. Removing auxiliary supervision leads to large performance drops on all three datasets. This fact demonstrates that the losses of dense prediction backbone serve as critical supervision roles during the training process. It further verifies that the supervision of V also constrains the predicted values of the implicit field, helping DPFs learn reasonable prediction results. In addition, auxiliary supervision on baseline prediction also benefits latent code z to learn high-level visual features.

\textbf{Effects of guidance image}. We conduct experiments to explore the effect of guidance images. Specifically, we train DPF models without guidance encoder and guidance latent code g on three datasets. The formulation of this simplified DPF is represented as:
\begin{align}
 v_x = f_\theta(z,x)
\end{align}
and the results are shown in Tab. \ref{tab:guide}. It's clear that DPFs with guidance get superior performance. Specifically, for PASCALContext, the mIoU of semantic segmentation is increased by 1.2\%, ADE20K performance increases by 1.3\%, and WHDR of IIW is  decreased by 0.6\%.  This indicates that guidance images can benefit the learning of interpolation parameters. We believe this plays a similar role to the guidance image in the guided image filter,  helping the learning of interpolation parameters, which makes the dense prediction results more consistent. Besides, CRFs like \cite{krahenbuhl2011efficient,arnab2016higher} are conventional techniques that work in the same spirit and we provide a comparison in the supplementary.

\textbf{Resolution of guidance image}. We also conduct experiments on the resolutions of guidance images, and the results are presented in Tab. \ref{tab:resolution}. As shown in the table, the results of the DPF with guidance always outperform the baseline model. Specifically, while the mIoU of baseline on Pascal dropped a lot with a 128$\times$128 input image, the DPF with 512$\times$512 guidance image improves the performance by 18.6\%.  Meanwhile,  when the resolution of the input image is the same, the larger the resolution of the guidance image, the better the performance of the model, which has been verified on all three datasets. This further illustrates the importance of guidance, while providing an appealing paradigm that trains low-resolution inputs with high-resolution guidance images.

\textbf{Visualization of latent code}. Fig. \ref{fig:tsne} presents visualizations of latent codes on three datasets. We use t-SNE to reduce the dimension of latent codes g and z to one and visualize them, respectively. As illustrated in the figure, latent code z encodes high-level features with semantic information, while latent code g focuses on low-level features with clear boundaries. Furthermore, latent code g preserves the details of the original image, but is relatively smoother, which benefits the learning of consistent DPFs.


\section{Conclusion}

In this paper, we propose dense prediction fields (DPFs), a new paradigm that makes dense value predictions for point coordinate queries. We use an implicit neural function to model the DPFs, which are compatible with point-level supervision. We verify the effectiveness of DPFs using two different tasks: semantic parsing and intrinsic image decomposition. We benchmark DPFs on three datasets including PASCALContext, ADE20K and IIW, and achieve state-of-the-art performance on all three datasets. 

{\small
\bibliographystyle{ieee_fullname}
\bibliography{egbib}

\begin{thebibliography}{10}\itemsep=-1pt

\bibitem{arbelaez2010contour}
Pablo Arbelaez, Michael Maire, Charless Fowlkes, and Jitendra Malik.
\newblock Contour detection and hierarchical image segmentation.
\newblock {\em IEEE transactions on pattern analysis and machine intelligence},
  33(5):898--916, 2010.

\bibitem{arnab2016higher}
Anurag Arnab, Sadeep Jayasumana, Shuai Zheng, and Philip~HS Torr.
\newblock Higher order conditional random fields in deep neural networks.
\newblock In {\em Computer Vision--ECCV 2016: 14th European Conference,
  Amsterdam, The Netherlands, October 11-14, 2016, Proceedings, Part II 14},
  pages 524--540. Springer, 2016.

\bibitem{baslamisli2021shadingnet}
Anil~S Baslamisli, Partha Das, Hoang-An Le, Sezer Karaoglu, and Theo Gevers.
\newblock Shadingnet: image intrinsics by fine-grained shading decomposition.
\newblock {\em International Journal of Computer Vision}, 129(8):2445--2473,
  2021.

\bibitem{baslamisli2018joint}
Anil~S Baslamisli, Thomas~T Groenestege, Partha Das, Hoang-An Le, Sezer
  Karaoglu, and Theo Gevers.
\newblock Joint learning of intrinsic images and semantic segmentation.
\newblock In {\em Proceedings of the European Conference on Computer Vision
  (ECCV)}, pages 286--302, 2018.

\bibitem{baslamisli2018cnn}
Anil~S Baslamisli, Hoang-An Le, and Theo Gevers.
\newblock Cnn based learning using reflection and retinex models for intrinsic
  image decomposition.
\newblock In {\em Proceedings of the IEEE conference on computer vision and
  pattern recognition}, pages 6674--6683, 2018.

\bibitem{bearman2016s}
Amy Bearman, Olga Russakovsky, Vittorio Ferrari, and Li Fei-Fei.
\newblock What’s the point: Semantic segmentation with point supervision.
\newblock In {\em European conference on computer vision}, pages 549--565.
  Springer, 2016.

\bibitem{bell2014intrinsic}
Sean Bell, Kavita Bala, and Noah Snavely.
\newblock Intrinsic images in the wild.
\newblock {\em ACM Transactions on Graphics (TOG)}, 33(4):1--12, 2014.

\bibitem{bi20151}
Sai Bi, Xiaoguang Han, and Yizhou Yu.
\newblock An l 1 image transform for edge-preserving smoothing and scene-level
  intrinsic decomposition.
\newblock {\em ACM Transactions on Graphics (TOG)}, 34(4):1--12, 2015.

\bibitem{chen2013simple}
Qifeng Chen and Vladlen Koltun.
\newblock A simple model for intrinsic image decomposition with depth cues.
\newblock In {\em Proceedings of the IEEE international conference on computer
  vision}, pages 241--248, 2013.

\bibitem{chen2022cerberus}
Xiaoxue Chen, Tianyu Liu, Hao Zhao, Guyue Zhou, and Ya-Qin Zhang.
\newblock Cerberus transformer: Joint semantic, affordance and attribute
  parsing.
\newblock In {\em Proceedings of the IEEE/CVF Conference on Computer Vision and
  Pattern Recognition}, pages 19649--19658, 2022.

\bibitem{chen2022pq}
Xiaoxue Chen, Hao Zhao, Guyue Zhou, and Ya-Qin Zhang.
\newblock Pq-transformer: Jointly parsing 3d objects and layouts from point
  clouds.
\newblock {\em IEEE Robotics and Automation Letters}, 7(2):2519--2526, 2022.

\bibitem{chen2021learning}
Yinbo Chen, Sifei Liu, and Xiaolong Wang.
\newblock Learning continuous image representation with local implicit image
  function.
\newblock In {\em Proceedings of the IEEE/CVF conference on computer vision and
  pattern recognition}, pages 8628--8638, 2021.

\bibitem{cordts2016cityscapes}
Marius Cordts, Mohamed Omran, Sebastian Ramos, Timo Rehfeld, Markus Enzweiler,
  Rodrigo Benenson, Uwe Franke, Stefan Roth, and Bernt Schiele.
\newblock The cityscapes dataset for semantic urban scene understanding.
\newblock In {\em Proceedings of the IEEE conference on computer vision and
  pattern recognition}, pages 3213--3223, 2016.

\bibitem{das2022intrinsic}
Partha Das, Sezer Karaoglu, and Theo Gevers.
\newblock Intrinsic image decomposition using physics-based cues and cnns.
\newblock {\em Computer Vision and Image Understanding}, 223:103538, 2022.

\bibitem{das2022pie}
Partha Das, Sezer Karaoglu, and Theo Gevers.
\newblock Pie-net: Photometric invariant edge guided network for intrinsic
  image decomposition.
\newblock In {\em Proceedings of the IEEE/CVF Conference on Computer Vision and
  Pattern Recognition}, pages 19790--19799, 2022.

\bibitem{degol2016geometry}
Joseph DeGol, Mani Golparvar-Fard, and Derek Hoiem.
\newblock Geometry-informed material recognition.
\newblock In {\em Proceedings of the IEEE conference on computer vision and
  pattern recognition}, pages 1554--1562, 2016.

\bibitem{dosovitskiy2020image}
Alexey Dosovitskiy, Lucas Beyer, Alexander Kolesnikov, Dirk Weissenborn,
  Xiaohua Zhai, Thomas Unterthiner, Mostafa Dehghani, Matthias Minderer, Georg
  Heigold, Sylvain Gelly, et~al.
\newblock An image is worth 16x16 words: Transformers for image recognition at
  scale.
\newblock {\em arXiv preprint arXiv:2010.11929}, 2020.

\bibitem{fan2018revisiting}
Qingnan Fan, Jiaolong Yang, Gang Hua, Baoquan Chen, and David Wipf.
\newblock Revisiting deep intrinsic image decompositions.
\newblock In {\em Proceedings of the IEEE conference on computer vision and
  pattern recognition}, pages 8944--8952, 2018.

\bibitem{forsyth2021intrinsic}
David Forsyth and Jason~J Rock.
\newblock Intrinsic image decomposition using paradigms.
\newblock {\em IEEE transactions on pattern analysis and machine intelligence},
  44(11):7624--7637, 2021.

\bibitem{gao2023semi}
Huan-ang Gao, Beiwen Tian, Pengfei Li, Xiaoxue Chen, Hao Zhao, Guyue Zhou,
  Yurong Chen, and Hongbin Zha.
\newblock From semi-supervised to omni-supervised room layout estimation using
  point clouds.
\newblock {\em arXiv preprint arXiv:2301.13865}, 2023.

\bibitem{gardner2019deep}
Marc-Andr{\'e} Gardner, Yannick Hold-Geoffroy, Kalyan Sunkavalli, Christian
  Gagn{\'e}, and Jean-Fran{\c{c}}ois Lalonde.
\newblock Deep parametric indoor lighting estimation.
\newblock In {\em Proceedings of the IEEE/CVF International Conference on
  Computer Vision}, pages 7175--7183, 2019.

\bibitem{genova2019learning}
Kyle Genova, Forrester Cole, Daniel Vlasic, Aaron Sarna, William~T Freeman, and
  Thomas Funkhouser.
\newblock Learning shape templates with structured implicit functions.
\newblock In {\em Proceedings of the IEEE/CVF International Conference on
  Computer Vision}, pages 7154--7164, 2019.

\bibitem{grosse2009ground}
Roger Grosse, Micah~K Johnson, Edward~H Adelson, and William~T Freeman.
\newblock Ground truth dataset and baseline evaluations for intrinsic image
  algorithms.
\newblock In {\em 2009 IEEE 12th International Conference on Computer Vision},
  pages 2335--2342. IEEE, 2009.

\bibitem{gupta2015indoor}
Saurabh Gupta, Pablo Arbel{\'a}ez, Ross Girshick, and Jitendra Malik.
\newblock Indoor scene understanding with rgb-d images: Bottom-up segmentation,
  object detection and semantic segmentation.
\newblock {\em International Journal of Computer Vision}, 112(2):133--149,
  2015.

\bibitem{he2012guided}
Kaiming He, Jian Sun, and Xiaoou Tang.
\newblock Guided image filtering.
\newblock {\em IEEE transactions on pattern analysis and machine intelligence},
  35(6):1397--1409, 2012.

\bibitem{hu2022learning}
Hanzhe Hu, Yinbo Chen, Jiarui Xu, Shubhankar Borse, Hong Cai, Fatih Porikli,
  and Xiaolong Wang.
\newblock Learning implicit feature alignment function for semantic
  segmentation.
\newblock {\em arXiv preprint arXiv:2206.08655}, 2022.

\bibitem{huang2018holistic}
Siyuan Huang, Siyuan Qi, Yixin Zhu, Yinxue Xiao, Yuanlu Xu, and Song-Chun Zhu.
\newblock Holistic 3d scene parsing and reconstruction from a single rgb image.
\newblock In {\em Proceedings of the European conference on computer vision
  (ECCV)}, pages 187--203, 2018.

\bibitem{janner2017self}
Michael Janner, Jiajun Wu, Tejas~D Kulkarni, Ilker Yildirim, and Josh
  Tenenbaum.
\newblock Self-supervised intrinsic image decomposition.
\newblock {\em Advances in neural information processing systems}, 30, 2017.

\bibitem{ji2017surfacenet}
Mengqi Ji, Juergen Gall, Haitian Zheng, Yebin Liu, and Lu Fang.
\newblock Surfacenet: An end-to-end 3d neural network for multiview stereopsis.
\newblock In {\em Proceedings of the IEEE International Conference on Computer
  Vision}, pages 2307--2315, 2017.

\bibitem{karsch2011rendering}
Kevin Karsch, Varsha Hedau, David Forsyth, and Derek Hoiem.
\newblock Rendering synthetic objects into legacy photographs.
\newblock {\em ACM Transactions on Graphics (TOG)}, 30(6):1--12, 2011.

\bibitem{krahenbuhl2011efficient}
Philipp Kr{\"a}henb{\"u}hl and Vladlen Koltun.
\newblock Efficient inference in fully connected crfs with gaussian edge
  potentials.
\newblock {\em Advances in neural information processing systems}, 24, 2011.

\bibitem{li2023lode}
Pengfei Li, Ruowen Zhao, Yongliang Shi, Hao Zhao, Jirui Yuan, Guyue Zhou, and
  Ya-Qin Zhang.
\newblock Lode: Locally conditioned eikonal implicit scene completion from
  sparse lidar.
\newblock {\em arXiv preprint arXiv:2302.14052}, 2023.

\bibitem{li2020inverse}
Zhengqin Li, Mohammad Shafiei, Ravi Ramamoorthi, Kalyan Sunkavalli, and
  Manmohan Chandraker.
\newblock Inverse rendering for complex indoor scenes: Shape, spatially-varying
  lighting and svbrdf from a single image.
\newblock In {\em Proceedings of the IEEE/CVF Conference on Computer Vision and
  Pattern Recognition}, pages 2475--2484, 2020.

\bibitem{li2018cgintrinsics}
Zhengqi Li and Noah Snavely.
\newblock Cgintrinsics: Better intrinsic image decomposition through
  physically-based rendering.
\newblock In {\em Proceedings of the European conference on computer vision
  (ECCV)}, pages 371--387, 2018.

\bibitem{li2020openrooms}
Zhengqin Li, Ting-Wei Yu, Shen Sang, Sarah Wang, Meng Song, Yuhan Liu, Yu-Ying
  Yeh, Rui Zhu, Nitesh Gundavarapu, Jia Shi, et~al.
\newblock Openrooms: An end-to-end open framework for photorealistic indoor
  scene datasets.
\newblock {\em arXiv preprint arXiv:2007.12868}, 2020.

\bibitem{lim2017enhanced}
Bee Lim, Sanghyun Son, Heewon Kim, Seungjun Nah, and Kyoung Mu~Lee.
\newblock Enhanced deep residual networks for single image super-resolution.
\newblock In {\em Proceedings of the IEEE conference on computer vision and
  pattern recognition workshops}, pages 136--144, 2017.

\bibitem{liu2020unsupervised}
Yunfei Liu, Yu Li, Shaodi You, and Feng Lu.
\newblock Unsupervised learning for intrinsic image decomposition from a single
  image.
\newblock In {\em Proceedings of the IEEE/CVF Conference on Computer Vision and
  Pattern Recognition}, pages 3248--3257, 2020.

\bibitem{ma2018single}
Wei-Chiu Ma, Hang Chu, Bolei Zhou, Raquel Urtasun, and Antonio Torralba.
\newblock Single image intrinsic decomposition without a single intrinsic
  image.
\newblock In {\em Proceedings of the European conference on computer vision
  (ECCV)}, pages 201--217, 2018.

\bibitem{meka2018lime}
Abhimitra Meka, Maxim Maximov, Michael Zollhoefer, Avishek Chatterjee,
  Hans-Peter Seidel, Christian Richardt, and Christian Theobalt.
\newblock Lime: Live intrinsic material estimation.
\newblock In {\em Proceedings of the IEEE conference on computer vision and
  pattern recognition}, pages 6315--6324, 2018.

\bibitem{michalkiewicz2019implicit}
Mateusz Michalkiewicz, Jhony~K Pontes, Dominic Jack, Mahsa Baktashmotlagh, and
  Anders Eriksson.
\newblock Implicit surface representations as layers in neural networks.
\newblock In {\em Proceedings of the IEEE/CVF International Conference on
  Computer Vision}, pages 4743--4752, 2019.

\bibitem{mildenhall2020nerf}
Ben Mildenhall, Pratul~P Srinivasan, Matthew Tancik, Jonathan~T Barron, Ravi
  Ramamoorthi, and Ren Ng.
\newblock Nerf: Representing scenes as neural radiance fields for view
  synthesis.
\newblock In {\em European conference on computer vision}, pages 405--421.
  Springer, 2020.

\bibitem{mottaghi2014role}
Roozbeh Mottaghi, Xianjie Chen, Xiaobai Liu, Nam-Gyu Cho, Seong-Whan Lee, Sanja
  Fidler, Raquel Urtasun, and Alan Yuille.
\newblock The role of context for object detection and semantic segmentation in
  the wild.
\newblock In {\em Proceedings of the IEEE conference on computer vision and
  pattern recognition}, pages 891--898, 2014.

\bibitem{narihira2015direct}
Takuya Narihira, Michael Maire, and Stella~X Yu.
\newblock Direct intrinsics: Learning albedo-shading decomposition by
  convolutional regression.
\newblock In {\em Proceedings of the IEEE international conference on computer
  vision}, pages 2992--2992, 2015.

\bibitem{narihira2015learning}
Takuya Narihira, Michael Maire, and Stella~X Yu.
\newblock Learning lightness from human judgement on relative reflectance.
\newblock In {\em Proceedings of the IEEE conference on computer vision and
  pattern recognition}, pages 2965--2973, 2015.

\bibitem{nestmeyer2017reflectance}
Thomas Nestmeyer and Peter~V Gehler.
\newblock Reflectance adaptive filtering improves intrinsic image estimation.
\newblock In {\em Proceedings of the IEEE conference on computer vision and
  pattern recognition}, pages 6789--6798, 2017.

\bibitem{park2019deepsdf}
Jeong~Joon Park, Peter Florence, Julian Straub, Richard Newcombe, and Steven
  Lovegrove.
\newblock Deepsdf: Learning continuous signed distance functions for shape
  representation.
\newblock In {\em Proceedings of the IEEE/CVF conference on computer vision and
  pattern recognition}, pages 165--174, 2019.

\bibitem{qian2019weakly}
Rui Qian, Yunchao Wei, Honghui Shi, Jiachen Li, Jiaying Liu, and Thomas Huang.
\newblock Weakly supervised scene parsing with point-based distance metric
  learning.
\newblock In {\em Proceedings of the AAAI Conference on Artificial
  Intelligence}, volume~33, pages 8843--8850, 2019.

\bibitem{ranftl2021vision}
Ren{\'e} Ranftl, Alexey Bochkovskiy, and Vladlen Koltun.
\newblock Vision transformers for dense prediction.
\newblock In {\em Proceedings of the IEEE/CVF International Conference on
  Computer Vision}, pages 12179--12188, 2021.

\bibitem{saxena2008make3d}
Ashutosh Saxena, Min Sun, and Andrew~Y Ng.
\newblock Make3d: Learning 3d scene structure from a single still image.
\newblock {\em IEEE transactions on pattern analysis and machine intelligence},
  31(5):824--840, 2008.

\bibitem{sengupta2019neural}
Soumyadip Sengupta, Jinwei Gu, Kihwan Kim, Guilin Liu, David~W Jacobs, and Jan
  Kautz.
\newblock Neural inverse rendering of an indoor scene from a single image.
\newblock In {\em Proceedings of the IEEE/CVF International Conference on
  Computer Vision}, pages 8598--8607, 2019.

\bibitem{shen2008intrinsic}
Li Shen, Ping Tan, and Stephen Lin.
\newblock Intrinsic image decomposition with non-local texture cues.
\newblock In {\em 2008 IEEE Conference on Computer Vision and Pattern
  Recognition}, pages 1--7. IEEE, 2008.

\bibitem{shen2011intrinsic}
Li Shen and Chuohao Yeo.
\newblock Intrinsic images decomposition using a local and global sparse
  representation of reflectance.
\newblock In {\em CVPR 2011}, pages 697--704. IEEE, 2011.

\bibitem{shen2013intrinsic}
Li Shen, Chuohao Yeo, and Binh-Son Hua.
\newblock Intrinsic image decomposition using a sparse representation of
  reflectance.
\newblock {\em IEEE transactions on pattern analysis and machine intelligence},
  35(12):2904--2915, 2013.

\bibitem{sitzmann2020implicit}
Vincent Sitzmann, Julien Martel, Alexander Bergman, David Lindell, and Gordon
  Wetzstein.
\newblock Implicit neural representations with periodic activation functions.
\newblock {\em Advances in Neural Information Processing Systems},
  33:7462--7473, 2020.

\bibitem{song2019neural}
Shuran Song and Thomas Funkhouser.
\newblock Neural illumination: Lighting prediction for indoor environments.
\newblock In {\em Proceedings of the IEEE/CVF Conference on Computer Vision and
  Pattern Recognition}, pages 6918--6926, 2019.

\bibitem{tang2021joint}
Jiaxiang Tang, Xiaokang Chen, and Gang Zeng.
\newblock Joint implicit image function for guided depth super-resolution.
\newblock In {\em Proceedings of the 29th ACM International Conference on
  Multimedia}, pages 4390--4399, 2021.

\bibitem{tian2022vibus}
Beiwen Tian, Liyi Luo, Hao Zhao, and Guyue Zhou.
\newblock Vibus: Data-efficient 3d scene parsing with viewpoint bottleneck and
  uncertainty-spectrum modeling.
\newblock {\em ISPRS Journal of Photogrammetry and Remote Sensing},
  194:302--318, 2022.

\bibitem{wei2017object}
Yunchao Wei, Jiashi Feng, Xiaodan Liang, Ming-Ming Cheng, Yao Zhao, and
  Shuicheng Yan.
\newblock Object region mining with adversarial erasing: A simple
  classification to semantic segmentation approach.
\newblock In {\em Proceedings of the IEEE conference on computer vision and
  pattern recognition}, pages 1568--1576, 2017.

\bibitem{wei2016stc}
Yunchao Wei, Xiaodan Liang, Yunpeng Chen, Xiaohui Shen, Ming-Ming Cheng, Jiashi
  Feng, Yao Zhao, and Shuicheng Yan.
\newblock Stc: A simple to complex framework for weakly-supervised semantic
  segmentation.
\newblock {\em IEEE transactions on pattern analysis and machine intelligence},
  39(11):2314--2320, 2016.

\bibitem{wu2019fastfcn}
Huikai Wu, Junge Zhang, Kaiqi Huang, Kongming Liang, and Yizhou Yu.
\newblock Fastfcn: Rethinking dilated convolution in the backbone for semantic
  segmentation.
\newblock {\em arXiv preprint arXiv:1903.11816}, 2019.

\bibitem{xu2019unstructuredfusion}
Lan Xu, Zhuo Su, Lei Han, Tao Yu, Yebin Liu, and Lu Fang.
\newblock Unstructuredfusion: Realtime 4d geometry and texture reconstruction
  using commercial rgbd cameras.
\newblock {\em IEEE transactions on pattern analysis and machine intelligence},
  42(10):2508--2522, 2019.

\bibitem{zhang2021unsupervised}
Qing Zhang, Jin Zhou, Lei Zhu, Wei Sun, Chunxia Xiao, and Wei-Shi Zheng.
\newblock Unsupervised intrinsic image decomposition using internal
  self-similarity cues.
\newblock {\em IEEE Transactions on Pattern Analysis and Machine Intelligence},
  2021.

\bibitem{zhao2017physics}
Hao Zhao, Ming Lu, Anbang Yao, Yiwen Guo, Yurong Chen, and Li Zhang.
\newblock Physics inspired optimization on semantic transfer features: An
  alternative method for room layout estimation.
\newblock In {\em Proceedings of the IEEE conference on computer vision and
  pattern recognition}, pages 10--18, 2017.

\bibitem{zhao2020pointly}
Hao Zhao, Ming Lu, Anbang Yao, Yiwen Guo, Yurong Chen, and Li Zhang.
\newblock Pointly-supervised scene parsing with uncertainty mixture.
\newblock {\em Computer Vision and Image Understanding}, 200:103040, 2020.

\bibitem{zheng2016deep}
Haitian Zheng, Lu Fang, Mengqi Ji, Matti Strese, Yigitcan {\"O}zer, and
  Eckehard Steinbach.
\newblock Deep learning for surface material classification using haptic and
  visual information.
\newblock {\em IEEE Transactions on Multimedia}, 18(12):2407--2416, 2016.

\bibitem{zheng2023steps}
Yupeng Zheng, Chengliang Zhong, Pengfei Li, Huan-ang Gao, Yuhang Zheng, Bu Jin,
  Ling Wang, Hao Zhao, Guyue Zhou, Qichao Zhang, et~al.
\newblock Steps: Joint self-supervised nighttime image enhancement and depth
  estimation.
\newblock {\em arXiv preprint arXiv:2302.01334}, 2023.

\bibitem{zhou2019semantic}
Bolei Zhou, Hang Zhao, Xavier Puig, Tete Xiao, Sanja Fidler, Adela Barriuso,
  and Antonio Torralba.
\newblock Semantic understanding of scenes through the ade20k dataset.
\newblock {\em International Journal of Computer Vision}, 127(3):302--321,
  2019.

\bibitem{zhou2019glosh}
Hao Zhou, Xiang Yu, and David~W Jacobs.
\newblock Glosh: Global-local spherical harmonics for intrinsic image
  decomposition.
\newblock In {\em Proceedings of the IEEE/CVF International Conference on
  Computer Vision}, pages 7820--7829, 2019.

\bibitem{zhou2015learning}
Tinghui Zhou, Philipp Krahenbuhl, and Alexei~A Efros.
\newblock Learning data-driven reflectance priors for intrinsic image
  decomposition.
\newblock In {\em Proceedings of the IEEE international conference on computer
  vision}, pages 3469--3477, 2015.

\bibitem{zhu2022irisformer}
Rui Zhu, Zhengqin Li, Janarbek Matai, Fatih Porikli, and Manmohan Chandraker.
\newblock Irisformer: Dense vision transformers for single-image inverse
  rendering in indoor scenes.
\newblock In {\em Proceedings of the IEEE/CVF Conference on Computer Vision and
  Pattern Recognition}, pages 2822--2831, 2022.

\bibitem{zoran2015learning}
Daniel Zoran, Phillip Isola, Dilip Krishnan, and William~T Freeman.
\newblock Learning ordinal relationships for mid-level vision.
\newblock In {\em Proceedings of the IEEE international conference on computer
  vision}, pages 388--396, 2015.

\end{thebibliography}
}

\newpage

\section{Appendix}

\subsection{Training Details}

We use the “poly” learning rate policy in which the current learning rate equals the base one multiplying $(1-\frac{epoch_{\rm current}}{epoch_{\rm max}})^{0.9}$. On PascalContext, we set the base learning rate to 0.028, while on ADE20k, we set the base learning rate to 0.035. For the IIW dataset, the base learning rate is set to 0.007. We set power to 0.9 on all datasets. Momentum and weight decay are set to 0.9 and 0.0001 respectively. The number of training epochs is set to 70 for PascalContext, 60 for ADE20k, and 30 for IIW dataset. For data augmentation, we exploit random rotation between -10 and 10 degrees and a random scale between 0.5 and 2 for all datasets. In addition, we add a random crop with the size of 512$\times$512 and a random horizontal flip. The comprehensive data augmentation scheme makes the network resist overfitting. Last, we normalize the RGB channels of images from a 0$\sim$255 range to a -1$\sim$1 range which can speed up the convergence of the model and improve the performance.

For PascalContext and ADE20k, we use \emph{DistributedDataParallel} and train DPFs with four GPUs with batch size 2 per GPU. For IIW, we train the model on a single GPU while setting the batch size to 2.

\subsection{Per-Category Evaluation}

\begin{figure*}[t]
  \centering
  \includegraphics[width=1\linewidth]{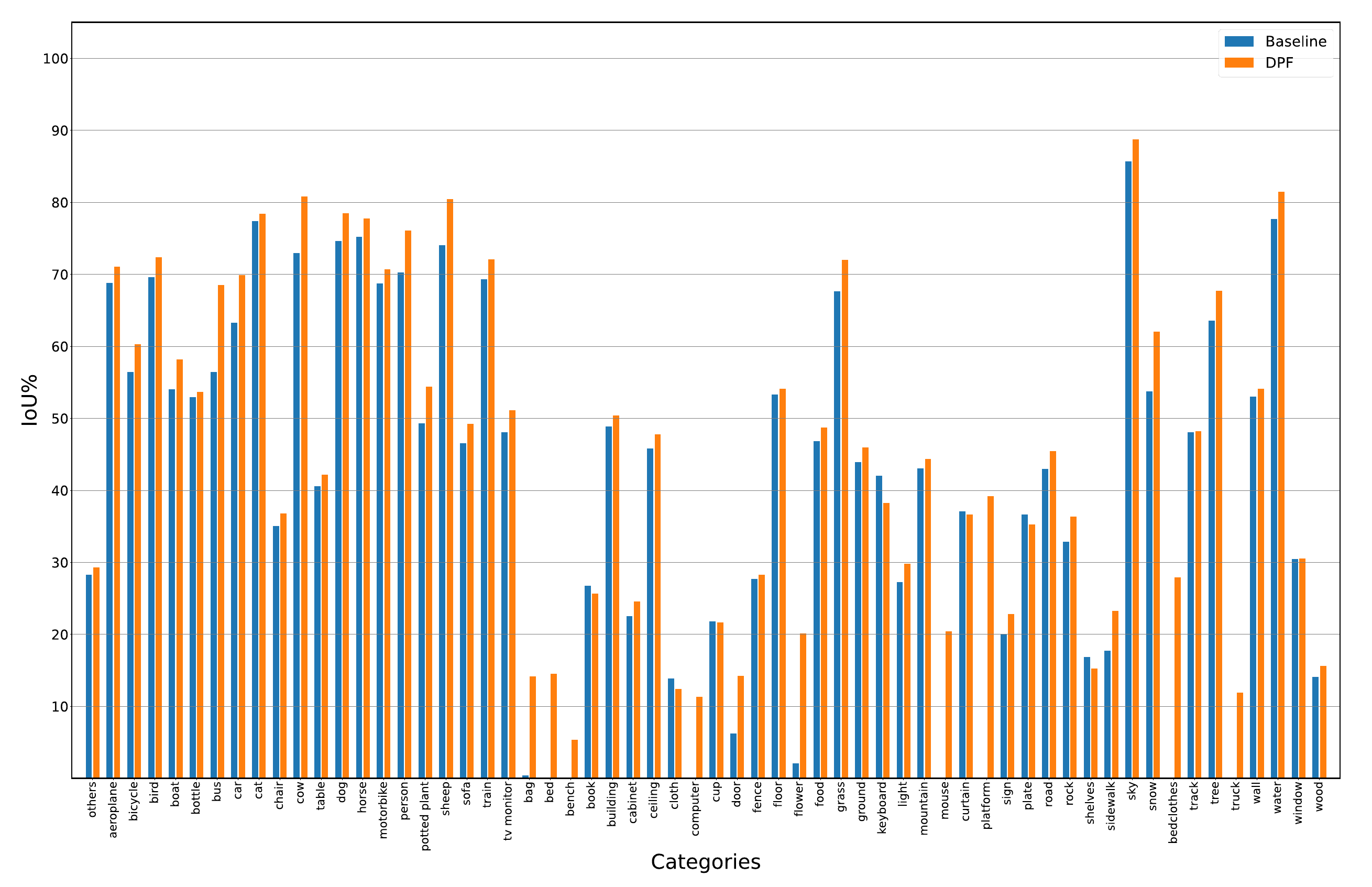}
  \caption{Per-category evaluation on PASCALContext dataset. }
  \label{fig:pascal_per_class} 
\end{figure*}

\begin{figure*}[t]
  \centering
  \includegraphics[width=1\linewidth]{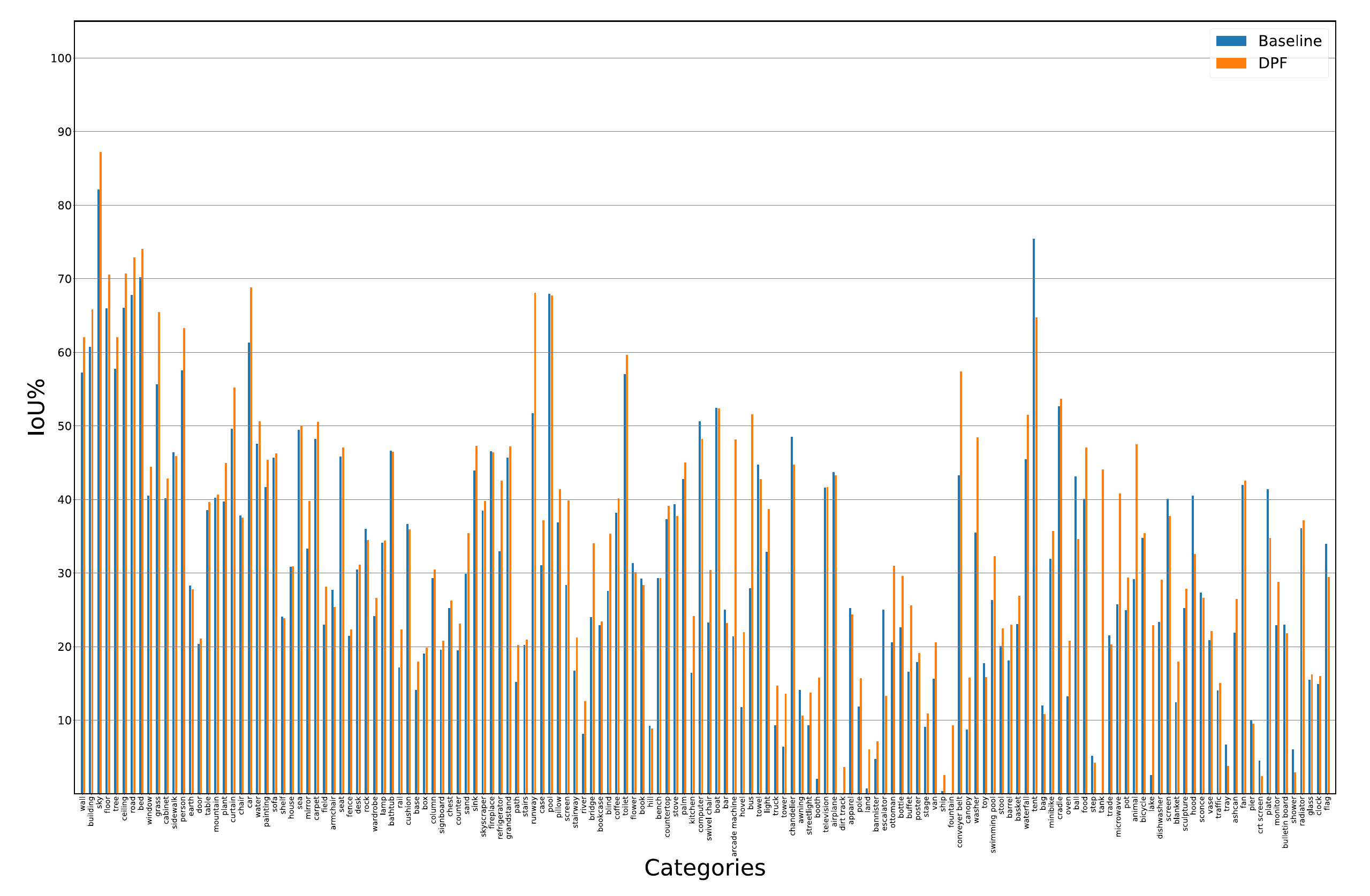}
  \caption{Per-category evaluation on ADE20k dataset. }
  \label{fig:ade_per_class} 
\end{figure*}

Fig.~\ref{fig:pascal_per_class}, \ref{fig:ade_per_class} show per-category semantic IoU  comparison between DPF and baseline on PASCALContext \cite{mottaghi2014role} and ADE20k \cite{zhou2019semantic} respectively. Tab. \ref{tab:pascal}, \ref{tab:ADE20k} provide specific IoU. It is manifested that our DPFs achieve better performance in most categories. Notably, the baseline results are 0 in some categories, (e.g., mouse, truck, and bus on PascalContext) while DPFs achieve significant growth of IoU in these categories. We believe this is credited to the guidance feature, which provides detailed semantic information and benefits the fine-grained segmentation.

\subsection{More Qualitative Results}

\begin{figure*}[t]
  \centering
  \includegraphics[width=1\linewidth]{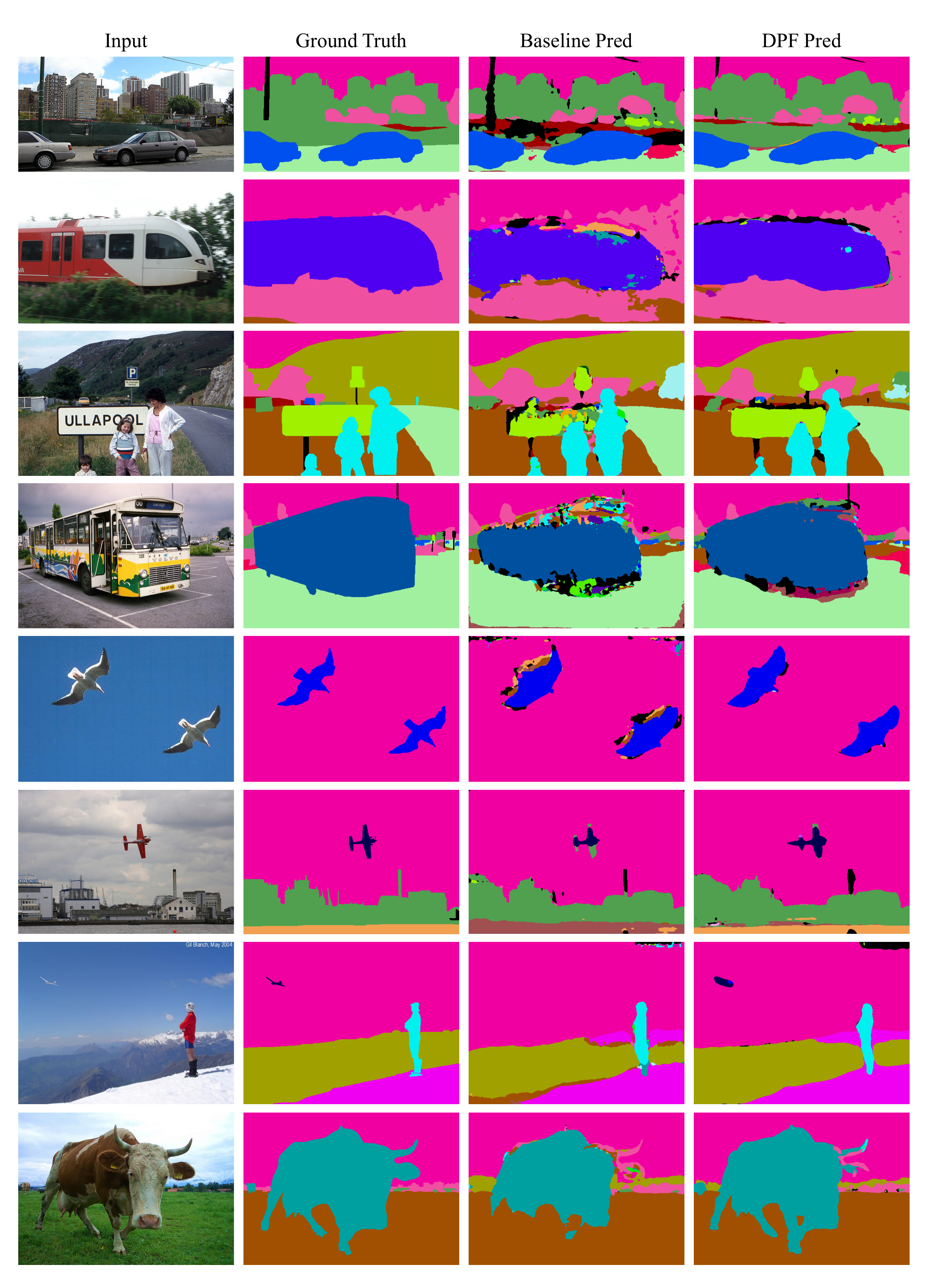}
  \caption{More qualitative results on PascalContext. }
  \label{fig:PASCALsup} 
\end{figure*}

\begin{figure*}[t]
  \centering
  \includegraphics[width=1\linewidth]{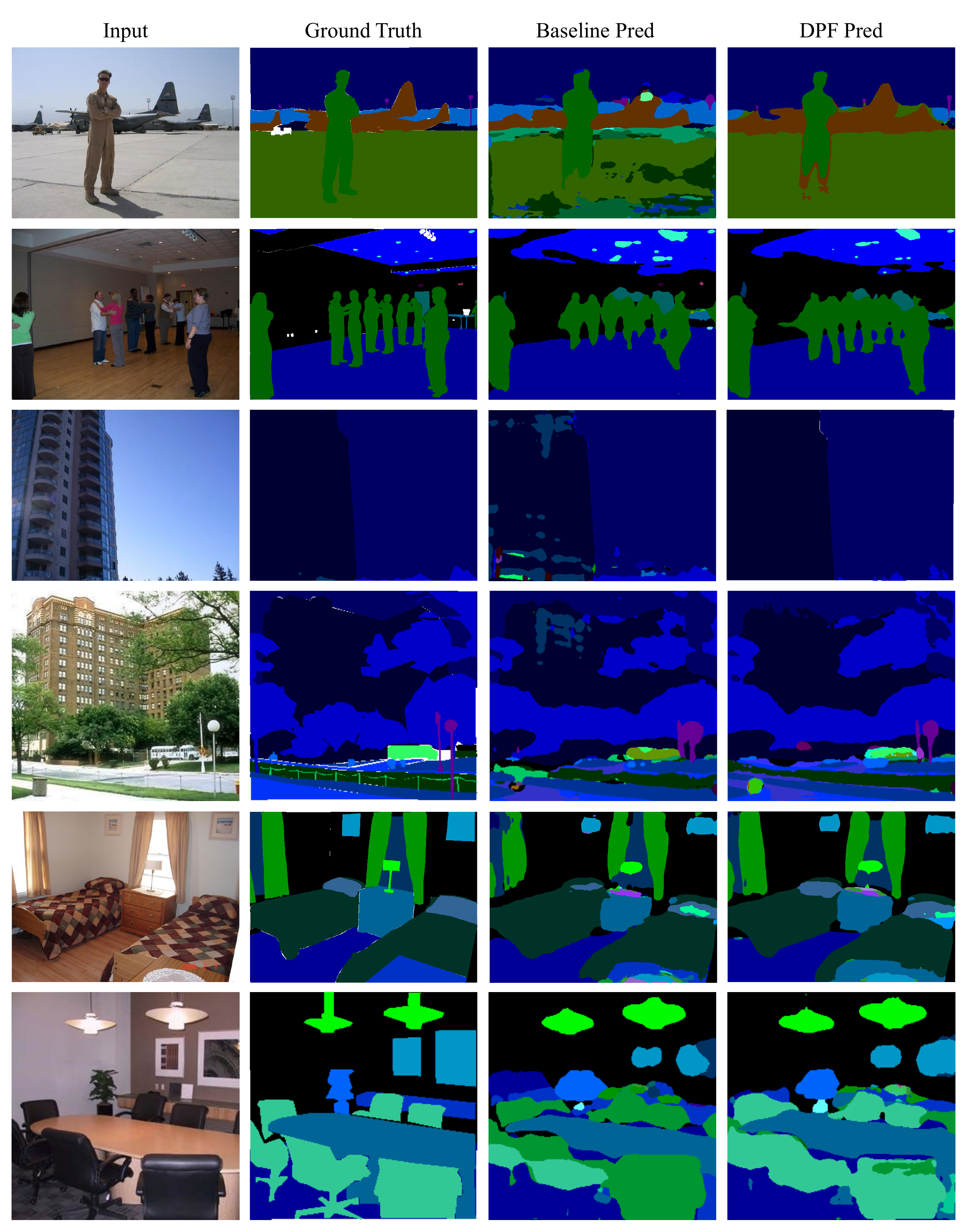}
  \caption{More qualitative results on ADE20k. White stands for the mask.}
  \label{fig:ADEsup} 
\end{figure*}

\textbf{Scene parsing}. Fig.\ref{fig:PASCALsup}, \ref{fig:ADEsup} shows some qualitative results for scene parsing on PascalContext and ADE20k datasets respectively. Compared with the baseline prediction, DPF produces more precise results. Specifically, the predictions of DPF have fewer noise patches, and the segmented shapes of objects are also more reasonable and clearer, which is credited to the smoothness of DPFs.


\textbf{Intrinsic decomposition}.  Fig. \ref{fig:iiw} presents some representative qualitative results on IIW. As shown in the area marked by the red box in the figure, DPF can distinguish the shadows of objects and successfully decompose the wall or ground into consistent reflectance. Also, it is manifested from the yellow box that our model can also predict the clothes with wrinkles as the same reflectance. These results illustrate the capability of DPF on intrinsic decomposition.

\subsection{More Ablation Study}

\begin{table}
	\centering
	\begin{tabular}{c|ccc}
		\toprule
		Weight & PASCAL & ADE20k & IIW \\
		\midrule
		Distance & 44.2 & 32.1 & 12.9  \\
        Learned  & \textbf{45.3} & \textbf{33.8} &  \textbf{11.9}  \\
		\bottomrule
	\end{tabular}
	\caption{Quantitative results on the options of weights.}
	\label{tab:weight}
\end{table}

\textbf{Interpolation weights}. We conduct an experiment to investigate the importance of using an MLP to predict the interpolation weights. We use bilinear interpolation with relative distances as weights, and the results are provided in Tab. \ref{tab:weight}. Obviously, the results of using the predicted weights are better, because the predicted weights not only leverage the spatial affinity between the query point and its neighbor pixels but also consider the semantic correlation provided by the latent code.

\begin{table}
	\centering
	\begin{tabular}{c|cc}
		\toprule
		Method & PASCAL & ADE20k  \\
		\midrule
        DPT & 40.4 & 30.4\\
		DPT+denseCRF & 41.3 & 31.1   \\
        DPT+DPF  & \textbf{45.3} & \textbf{33.8}  \\
		\bottomrule
	\end{tabular}
	\caption{Quantitative results on the effects of DPF.}
	\label{tab:crf}
\end{table}

\begin{table}[t]
	\centering
    \small
	\begin{tabular}{c|c|c}
		\toprule
		\textbf{Dataset} & \textbf{Method} & \textbf{mIoU} \\
		\midrule
		 \multirow{2}[0]{*}{PASCAL} & Baseline(\%) & 46.5\\ 
		& w/DPF (\%) & \textbf{61.2} (+14.7)\\
		\midrule
		\multirow{2}[0]{*}{ADE20k} & Baseline(\%) & 38.0\\ 
		& w/DPF (\%) & \textbf{42.4} (+4.4)\\
		\bottomrule
	\end{tabular}
	\caption{Transductive learning on PASCALContext and ADE20k unlabeled training sets, compared with DPT(baseline).}
	\label{tab:transductive}
\end{table}

\begin{table*}[t]
	\centering
	\scriptsize
	\begin{tabular}{c|ccccccccccc}
		\toprule
		\textbf{Method} & \textbf{Wood} & \textbf{Painted} & \textbf{Paper} & \textbf{Glass} & \textbf{Brick} & \textbf{Metal} & \textbf{Flat} & \textbf{Plastic} & \textbf{Textured} & \textbf{Glossy} & \textbf{Shiny}\\
		\midrule
		Baseline(\%) & 71.0 & 82.3 & 30.6 & 29.5 & 79.7 & 7.4 & 3.1 & 18.5 & 57.6 & 21.8 & 55.1\\ 
		w/DPF (\%) & 72.5(+1.5) & 84.0(+1.7) & 31.4(+0.8) & 30.5(+1.0) & 81.3(+0.6) & 7.9(+0.5) & 3.1(+0) & 18.9(+0.4) & 58.4(+0.8) & 22.2(+0.4) & 56.3(+1.2)\\
		\bottomrule
	\end{tabular}
	\caption{Precision of reflectance prediction on different attributes.}
	\label{tab:attribute}
\end{table*}

\textbf{Comparsions with CRF method}. CRFs are conventional techniques that smooth the dense prediction results on top of dense prediction baselines using low-level features, which work in a similar spirit to DPF. To compare with CRF, we additionally post-process the outputs of DPT baselines using denseCRF \cite{krahenbuhl2011efficient}, and the results are shown in Tab. \ref{tab:crf}. Since the prediction of intrinsic decomposition is a continuous value between 0-1, and the denseCRF method requires a discrete label set, we only provide the results on ADE20k and PASCAL. As shown in the table, DPF outperforms the denseCRF method with clear margins, demonstrating its superiority with CRF.

\subsection{Analysis Experiment}

\textbf{DPF is effective in the intrinsic decomposition of different materials.} Tab. \ref{tab:attribute} presents the reflectance precision on different materials. We use a pre-trained model of \cite{chen2022cerberus} to pre-process the IIW dataset and get attribute maps for each test image. For a pair of comparison points, we can easily get their attribute from the coordinates on the attribute maps. If the predicted relative reflectance of the comparison pair is correct, we consider the reflectance predictions of the two points to be correct. With the above preliminaries,  we calculate the per-attribute precision of both baseline and DPF. The results show that DPF has superior performance in different attributes compared with the baseline. Moreover, DPF gets higher precision on attributes such as wall and painted, while glass and metal have lower precision, which demonstrates that the model works better on diffuse materials, while intrinsic decomposition with specular reflection is relatively difficult.

\subsection{Transductive Learning}
\textbf{DPF achieves superior performance on both inductive learning and transductive learning.} Inductive learning, which represents the generalization of models, means learning from one sample and testing on another unseen sample. While transductive learning refers to testing on the unlabeled sample where we have obtained features during training. Commonly, weakly supervised learning may refer to both inductive learning and transductive learning because of the number of unlabeled pixels in the image. In Table. \ref{tab:transductive} , it is obvious that our DPF is also effective in transductive learning on both datasets. Especially on PascalContext, DPF has a strong lead of 14.7\%. This demonstrates that our proposed DPF is a general framework that is not specific to the inductive or transductive solution.

\begin{table*}
\scriptsize
  \centering
  \begin{tabular}{c|cccccccccc}
    \toprule
    \textbf{Method} &others&aeroplane&bicycle&bird&boat&bottle&bus&car&cut&chair  \\
    \midrule
    Baseline(\%) &28.3&68.8&56.4&69.6&54.0&52.9&56.5&63.3&77.4&35.1 \\
    DPF(\%) & \textbf{29.3}&\textbf{71.1}&\textbf{60.3}&\textbf{72.4}&\textbf{58.2}&\textbf{53.7}&\textbf{68.5}&\textbf{69.9}&\textbf{78.4}&\textbf{36.8} \\
    \toprule
    \textbf{Method} &cow&table&dog&horse&motorbike&person&potted plant&sheep&sofa&train \\
    \midrule
    Baseline(\%) & 72.9&40.6&74.6&75.2&68.7&70.3&49.3&74.1&46.5&69.3\\
    DPF(\%) & \textbf{80.8}&\textbf{42.2}&\textbf{78.5}&\textbf{77.8}&\textbf{70.7}&\textbf{76.1}&\textbf{54.4}&\textbf{80.5}&\textbf{49.2}&\textbf{72.1}\\
    \toprule
    \textbf{Method} & TV monitor & bag & bed & bench & book & building & cabinet & ceiling & cloth & computer  \\
    \midrule
    Baseline(\%)&48.1&0.4&0.0&0.0&\textbf{26.7}&48.9&22.6&45.8&\textbf{13.9}&0.0 \\
    DPF(\%) & \textbf{51.1}&\textbf{14.1}&\textbf{14.5}&\textbf{5.4}&25.7&\textbf{50.4}&\textbf{24.6}&\textbf{47.8}&12.4&\textbf{11.3}\\
    \toprule
    \textbf{Method} & cup & door & fence & floor & flower & food & grass & ground & keyboard & light  \\
    \midrule
    Baseline(\%) &\textbf{21.8}&6.2&27.7&53.3&2.1&46.9&67.6&43.9&\textbf{42.0}&27.3 \\
    DPF(\%) & 21.7&\textbf{14.3}&\textbf{28.3}&\textbf{54.1}&\textbf{20.1}&\textbf{48.7}&\textbf{72.0}&\textbf{45.9}&38.3&\textbf{29.8}\\
    \toprule
    \textbf{Method} & mountain & mouse & curtain & platform & sign & plate & road & rock & shelves & sidewalk  \\
    \midrule
    Baseline(\%) &43.1&0.0&\textbf{37.1}&0.0&20.0&\textbf{36.6}&43.0&32.8&\textbf{16.8}&17.7 \\
    DPF(\%) & \textbf{44.4}&\textbf{20.4}&36.6&\textbf{39.2}&\textbf{22.8}&35.3&\textbf{45.4}&\textbf{36.3}&15.3&\textbf{23.3}\\
    \toprule
    \textbf{Method} & sky & snow & bedclothes & track & tree & truck & wall & water & window & wood  \\
    \midrule
    Baseline(\%) &85.7&53.7&0.1&48.1&63.6&0.0&53.0&77.7&30.4&14.1 \\
    DPF(\%) & \textbf{88.7}&\textbf{62.1}&\textbf{27.9}&\textbf{48.2}&\textbf{67.7}&\textbf{11.9}&\textbf{54.1}&\textbf{81.4}&\textbf{30.5}&\textbf{15.7}\\
    \bottomrule
  \end{tabular}
  \caption{Per-category semantic parsing results on PascalContext.}
  \label{tab:pascal}
\end{table*}

\begin{table*}
\scriptsize
  \centering
  \begin{tabular}{c|cccccccccc}
    \toprule
    \textbf{Method} &wall&building&sky&floor&tree&ceiling&road&bed&window&grass  \\
    \midrule
    Baseline(\%) &57.2&60.8&82.1&66.0&57.8&66.0&67.8&70.2&40.5&55.6\\    
    DPF(\%) &\textbf{62.0}&\textbf{65.8}&\textbf{87.2}&\textbf{70.5}&\textbf{62.0}&\textbf{70.7}&\textbf{72.9}&\textbf{74.0}&\textbf{44.4}&\textbf{65.5}\\
    \toprule
    \textbf{Method} &cabinet&sidewalk&person&earth&door&table&mountain&plant&curtain&chair\\
    \midrule
    Baseline(\%) &40.2&\textbf{46.4}&57.6&\textbf{28.3}&20.3&38.6&40.2&39.7&49.6&\textbf{37.8}\\
    DPF(\%) &\textbf{42.9}&45.9&\textbf{63.3}&27.8&\textbf{21.1}&\textbf{39.7}&\textbf{40.7}&\textbf{44.9}&\textbf{55.2}&37.5\\
    \toprule
    \textbf{Method} &car&water&painting&sofa&shelf&house&sea&mirror&carpet&field  \\
    \midrule
    Baseline(\%)&61.3&47.6&41.7&45.6&\textbf{24.1}&30.8&49.4&33.3&48.2&23.0 \\
    DPF(\%) &\textbf{68.8}&\textbf{50.6}&\textbf{45.4}&\textbf{46.2}&23.9&\textbf{30.9}&\textbf{50.0}&\textbf{39.8}&\textbf{50.5}&\textbf{28.1}\\
    \toprule
    \textbf{Method} &armchair&seat&fence&desk&rock&wardrobe&lamp&bathtub&rail&cushion  \\
    \midrule
    Baseline(\%) &\textbf{27.7}&45.8&21.4&30.4&\textbf{36.0}&24.2&34.1&\textbf{45.6}&17.2&\textbf{36.7} \\
    DPF(\%) &25.4&\textbf{47.1}&\textbf{22.3}&\textbf{31.2}&34.5&\textbf{26.6}&\textbf{34.4}&46.5&\textbf{22.3}&35.9\\
    \toprule
    \textbf{Method} &base&box&column&signboard&chest&counter&sand&sink&skyscraper&fireplace  \\
    \midrule
    Baseline(\%) &14.1&19.1&29.3&19.6&25.2&19.5&29.9&43.9&38.4&\textbf{46.5} \\
    DPF(\%) &\textbf{17.9}&\textbf{19.8}&\textbf{30.5}&\textbf{20.8}&\textbf{26.3}&\textbf{23.1}&\textbf{35.4}&\textbf{47.3}&\textbf{39.8}&46.4\\
    \toprule
    \textbf{Method} &refrigerator&grandstand&path&stairs&runway&case&pool&pillow&screen&stairway  \\
    \midrule
    Baseline(\%) &32.9&45.7&15.2&20.2&51.7&31.0&\textbf{68.0}&36.9&28.4&16.7 \\
    DPF(\%) &\textbf{42.6}&\textbf{47.2}&\textbf{20.2}&\textbf{20.9}&\textbf{68.1}&\textbf{37.1}&67.7&\textbf{41.4}&\textbf{39.8}&\textbf{21.2}\\
    \toprule
    \textbf{Method} &river&bridge&bookcase&blind&coffee&toilet&flower&book&hill&bench  \\
    \midrule
    Baseline(\%) &8.2&24.0&22.9&27.6&38.1&57.0&\textbf{31.3}&\textbf{29.2}&\textbf{9.2}&29.3 \\
    DPF(\%) &\textbf{12.6}&\textbf{34.0}&\textbf{23.4}&\textbf{35.3}&\textbf{40.1}&\textbf{59.6}&30.1&28.4&8.9&\textbf{29.3}\\
        \toprule
    \textbf{Method} &countertop&stove&palm&kitchen&computer&swivel chair&boat&bar&arcade machine&hovel  \\
    \midrule
    Baseline(\%) &37.3&\textbf{39.4}&42.8&16.5&\textbf{50.6}&23.2&\textbf{52.5}&\textbf{25.0}&21.3&11.8 \\
    DPF(\%) &\textbf{39.1}&37.7&\textbf{45.0}&\textbf{24.1}&48.2&\textbf{30.4}&52.4&23.2&\textbf{48.2}&\textbf{21.9}\\
    \toprule
    \textbf{Method} &bus&towel&light&truck&tower&chandelier&awning&streetlight&booth&television  \\
    \midrule
    Baseline(\%) &27.9&\textbf{44.8}&32.9&9.3&6.4&\textbf{48.5}&\textbf{14.1}&9.3&2.0&41.6 \\
    DPF(\%) &\textbf{51.6}&42.7&\textbf{38.7}&\textbf{14.7}&\textbf{13.6}&44.7&10.6&\textbf{13.8}&\textbf{15.7}&\textbf{41.6}\\
    \toprule
    \textbf{Method} &airplane&dirt track&apparel&pole&land&bannister&escalator&ottoman&bottle&buffet  \\
    \midrule
    Baseline(\%) &\textbf{43.7}&0.0&\textbf{25.2}&11.8&0.7&4.7&\textbf{25.0}&20.6&22.6&16.5 \\
    DPF(\%) &43.3&\textbf{3.7}&24.3&\textbf{15.7}&\textbf{6.0}&\textbf{7.1}&13.3&\textbf{31.0}&\textbf{29.6}&\textbf{25.6}\\
    \toprule
    \textbf{Method} &poster&stage&van&ship&fountain&conveyer belt&canopy&washer&toy&swimming pool  \\
    \midrule
    Baseline(\%) &17.9&9.1&15.6&0.3&0.0&43.2&8.7&35.5&\textbf{17.7}&26.3 \\
    DPF(\%) &\textbf{19.1}&\textbf{10.9}&\textbf{20.6}&\textbf{2.5}&\textbf{9.3}&\textbf{57.4}&\textbf{15.8}&\textbf{48.4}&15.8&\textbf{32.3}\\
    \toprule
    \textbf{Method} &stool&barrel&basket&waterfall&tent&bag&minibike&cradle&oven&ball  \\
    \midrule
    Baseline(\%) &20.1&18.1&23.1&45.4&\textbf{75.4}&\textbf{12.0}&31.9&52.7&13.2&\textbf{43.2} \\
    DPF(\%) &\textbf{22.5}&\textbf{23.0}&\textbf{26.9}&\textbf{51.5}&64.7&10.8&\textbf{35.7}&\textbf{53.6}&\textbf{20.8}&34.6\\
    \toprule
    \textbf{Method} &food&step&tank&trade&microwave&pot&animal&bicycle&lake&dishwasher  \\
    \midrule
    Baseline(\%) &40.1&\textbf{5.1}&0.0&\textbf{21.5}&25.7&24.9&29.2&34.8&2.5&23.3 \\
    DPF(\%) &\textbf{47.1}&4.2&\textbf{44.1}&20.2&\textbf{40.8}&\textbf{29.4}&\textbf{47.5}&\textbf{35.4}&\textbf{22.9}&\textbf{29.1}\\
    \toprule
    \textbf{Method} &screen&blanket&sculpture&hood&sconce&vase&traffic&tray&ashcan&fan  \\
    \midrule
    Baseline(\%) &\textbf{40.1}&12.5&25.2&\textbf{40.5}&\textbf{27.3}&20.8&14.1&\textbf{6.7}&21.9&42.0 \\
    DPF(\%) &37.7&\textbf{17.9}&\textbf{27.9}&32.6&26.6&\textbf{22.1}&\textbf{15.0}&3.8&\textbf{26.5}&\textbf{42.5}\\
    \toprule
    \textbf{Method} &pier&crt screen&plate&monitor&bulletin board&shower&radiator&glass&clock&flag  \\
    \midrule
    Baseline(\%) &\textbf{9.9}&\textbf{4.5}&\textbf{41.4}&22.9&\textbf{23.0}&\textbf{6.0}&36.1&15.5&14.9&\textbf{33.9} \\
    DPF(\%) &9.5&2.4&34.7&\textbf{28.8}&21.8&2.9&\textbf{37.2}&\textbf{16.2}&\textbf{16.0}&29.4\\
    \bottomrule
  \end{tabular}
  \caption{Per-category semantic parsing results on ADE20k.}
  \label{tab:ADE20k}
\end{table*}

\begin{figure*}[t]
  \centering
  \includegraphics[width=1\linewidth]{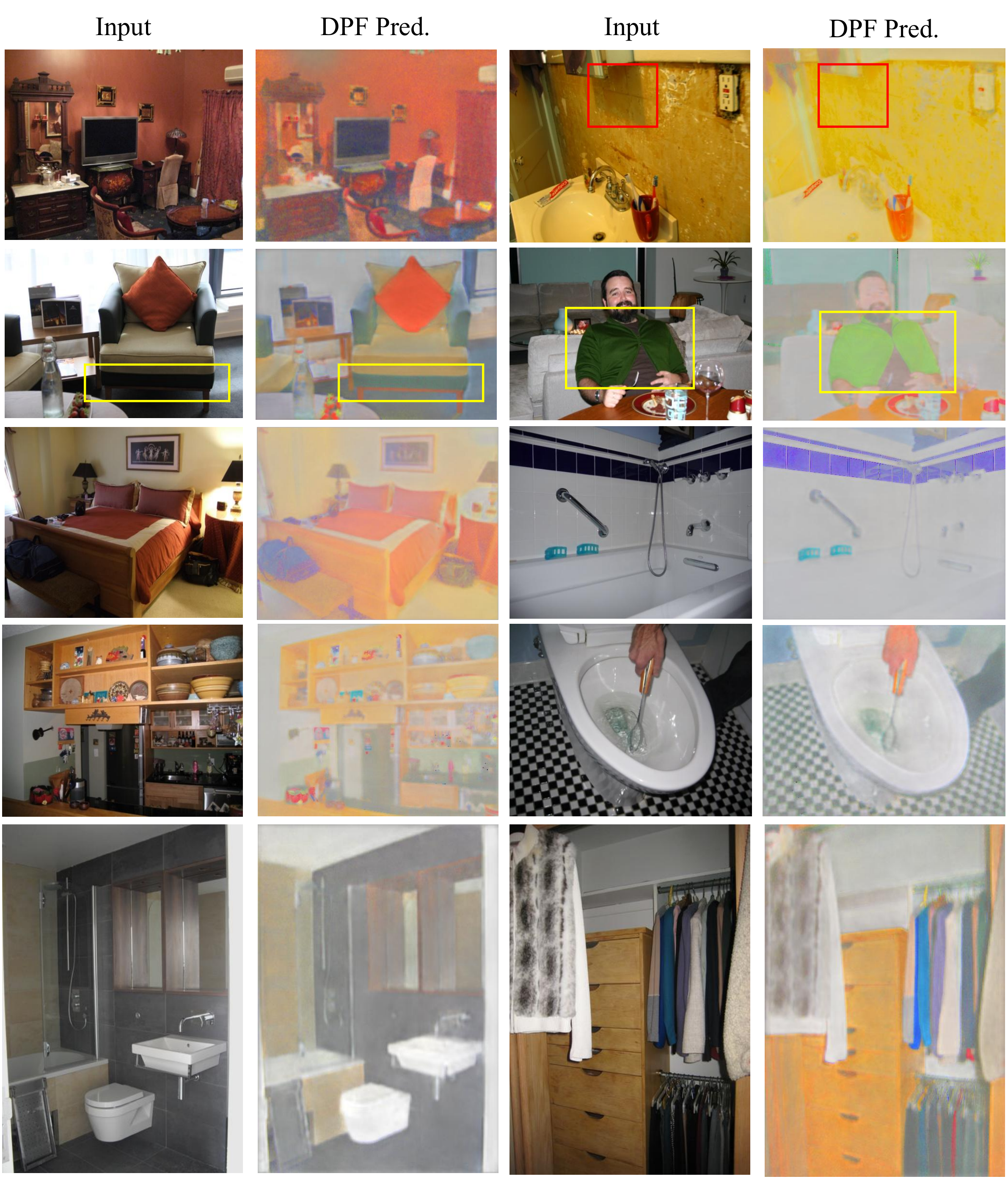}
  \caption{More qualitative results on IIW. }
  \label{fig:iiw} 
\end{figure*}

\end{document}